\documentclass[preprint,review,11pt]{elsarticle}
\usepackage{amssymb}
\usepackage{geometry}
\usepackage{amsmath}
\usepackage{wrapfig}
\usepackage{subfig}
\usepackage[linesnumbered,ruled]{algorithm2e}
\usepackage{multirow}
\usepackage{xcolor,graphicx,float}
\usepackage{paralist}
\usepackage{setspace}

\usepackage{fancyhdr}
\begin{document}

\begin{frontmatter}

\title{
\setlength{\baselineskip}{25pt}
DynamicKD: An Effective Knowledge Distillation via Dynamic Entropy Correction-Based Distillation for Gap Optimizing}

\author[author1]{Songling~Zhu}

\author[author1]{Ronghua~Shang\corref{cor1}}
\cortext[cor1]{Corresponding authors.}
\ead{rhshang@mail.xidian.edu.cn}

\author[author2]{Bo~Yuan}

\author[author1]{Weitong~Zhang}

\author[author1]{Yangyang~Li}

\author[author1]{Licheng~Jiao}

\address[author1]{\setlength{\baselineskip}{17pt}Key Laboratory of Intelligent Perception and Image Understanding of Ministry of Education, School of Artificial Intelligence, Xidian University, Xi’an, Shaanxi Province 710071, China.}
\address[author2]{\setlength{\baselineskip}{17pt}Guangdong Provincial Key Laboratory of Brain-inspired Intelligent Computation, Southern University of Science and Technology, Shenzhen 518055, China.
\vspace{-9mm}
}

\begin{abstract} 
The knowledge distillation uses a high-performance teacher network to guide the student network. However, the performance gap between the teacher and student networks can affect the student's training. This paper proposes a novel knowledge distillation algorithm based on dynamic entropy correction to reduce the gap by adjusting the student instead of the teacher. 
Firstly, the effect of changing the output entropy (short for output information entropy) in the student on the distillation loss is analyzed in theory. This paper shows that correcting the output entropy can reduce the gap. 
Then, a knowledge distillation algorithm based on dynamic entropy correction is created, which can correct the output entropy in real-time with an entropy controller updated dynamically by the distillation loss. 
The proposed algorithm is validated on the CIFAR100 and ImageNet. The comparison with various state-of-the-art distillation algorithms shows impressive results, especially in the experiment on the CIFAR100 regarding teacher-student pair resnet32x4-resnet8x4. The proposed algorithm raises 2.64 points over the traditional distillation algorithm and 0.87 points over the state-of-the-art algorithm CRD in classification accuracy, demonstrating its effectiveness and efficiency.
\end{abstract}

\vspace{10cm}


\end{frontmatter}

\section{Introduction}
Deep neural networks have been successfully applied to various computer vision tasks, 
such as 
few-shot learning \cite{peng_few-shot_2019},
semantic segmentation \cite{li_ctnet_2022},
image retrieval \cite{yu_learning_2015, li_deep_2019}, image click prediction \cite{yu_hierarchical_2022}, and Human pose processing \cite{hong_multimodal_2019, hong_multimodal_2015, hong_image-based_2015}.
To obtain high-performance neural networks, researchers have designed broader and deeper network structures (such as GoogLeNet \cite{szegedy_going_2015} and ResNet \cite{he_deep_2016}) and even utilized neural architecture search algorithms to develop more complex network topologies \cite{zhang_semi-supervised_2021, shang_evolutionary_2022}. These large-scale network structures can be trained and inferred on powerful workstations or GPUs, yet it is challenging to deploy them on resource-constrained devices, such as embedded or mobile devices \cite{sandler_mobilenetv2_2018}. Therefore, model compression and acceleration are of great research interest and value \cite{cheng_survey_2020}.

The famous model compression algorithms are model pruning \cite{yao_deep_2021} and knowledge distillation \cite{mirzadeh_improved_2020}. 
They can significantly reduce the model complexity and speed up model inference with acceptable accuracy loss. Model pruning can find and remove redundant structures in large-scale networks. These redundant structures contribute less to network performance
\cite{chen_shallowing_2019}. 
However, model pruning requires numerous iterations \cite{guo_dmcp_2020} and cumbersome fine-tuning operations \cite{he_asymptotic_2020}, making applying model pruning more difficult.
Knowledge distillation is an easy-to-use model compression method that uses a trained large-scale network to guide the training of a compact network \cite{hinton_distilling_2015}. Buciluǎ \textit{et al.} first mentioned this method to transfer knowledge from the ensemble of multiple models to another model \cite{bucilua_model_2006}. Hinton \textit{et al.} first introduced the concept of knowledge distillation by increasing the temperature so that the teacher network can generate logits (located before the last softmax layer) containing rich inter-classes similarity information \cite{hinton_distilling_2015}. This information is more valuable to guide the training of the student network than the ground truth labels. Without cumbersome redundancy measures and fine-tuning operations, knowledge distillation has been popular in many application fields, such as 
visual question answering \cite{10.1145/3487042},
text recognition \cite{wang_joint_2021},
Hyperspectral Image Classification \cite{shang_hyperspectral_2022},
and person search \cite{li_hierarchical_2021}. Therefore, this paper focuses on knowledge distillation-based model compression.

In the knowledge distillation, how the teacher network better guides the student network training has become vital research. 
Several researchers have studied various knowledge used in the distillation process. 
Romero \textit{et al.} applied the middle layer features of the teacher network as knowledge to guide the student network learning \cite{romero_fitnets_2015}. 
Komodakis \textit{et al.} transferred attention from the large-scale teacher network to the student network \cite{komodakis_paying_2017}. 
Damiano \textit{et al.} viewed the knowledge transfer between teacher and student networks as maximizing the mutual information between teacher and student networks \cite{ahn_variational_2019}. 
Zagoruyko \textit{et al.} used the attention mechanism as a learnable knowledge, and they used the teacher network's attention knowledge to guide the student network's training.
In neural networks, the convolutional layers map one feature to another. Yim \textit{et al.} treated the mapping processing of features between layers as knowledge and used the FSP matrix to describe this knowledge so that the student network could imitate it \cite{yim_gift_2017}. 
The rich and varied knowledge exchange between the teacher and student networks helps the student network training.
But, these methods do not focus on the performance gap between the teacher and student networks, which may affect student network learning.

Several works have studied the performance gap and proposed corresponding improvements. Cho \textit{et al.} found that the underperformance teacher with early-stop training benefits the student \cite{cho_efficacy_2019}. Mirzadeh \textit{et al.} found that when the gap between the teacher network and the student network is large, the student trained by a lower-performance lightweight teacher network performs better than the one taught by a higher-performance large-scale teacher network \cite{mirzadeh_improved_2020}. 
For this reason, he utilized a medium-sized neural network (called Teacher Assistant) to help the student network cross the large performance gap. 
Both methods mentioned above show that reducing the performance gap can improve distillation performance. 
However, these static methods do not correct the performance gap further during the distillation. 
The performance gap keeps changing with the performance improvement of the student network, so these strategies may still hinder the student network from imitating the high-performance teacher network.

Various knowledge distillation algorithms are available to continuously update the knowledge applied for the student network training during the distillation. 
In other words, these methods provide a way to change the student network's training difficulty dynamically. 
According to the source of knowledge, these algorithms can be divided into two categories. 
One generates different complex knowledge with the teacher network. Zhao \textit{et al.} trained the teacher and student networks together from scratch so that the knowledge difficulty could increase with the improvement of the student network performance \cite{zhao_highlight_2020}. Jin \textit{et al.} used multiple teacher networks with different performances to teach the student network sequentially during the distillation \cite{jin_knowledge_2019}. The other makes multiple teacher networks teach each other or collaborate to generate a more potent teacher network so that the knowledge difficulty can rise as distillation proceeds. DML uses multiple student networks, and each student network learns knowledge from other students \cite{zhang_deep_2018}. ONE constructs a robust teacher network from multiple student networks by a learnable gate component \cite{lan_knowledge_2018}. KDCL investigates multiple knowledge ensemble methods to generate high-quality knowledge from many student networks \cite{guo_online_2020}. PCL utilizes multiple student networks to develop the Meam Teacher and the ensemble teacher network to guide the training of these student networks \cite{wu_peer_2021}. However, the teacher networks with different performances vary the knowledge quality, which may mislead the student network training.
The same phenomenon was also found in the experimental analysis of this paper.
In addition, their training and storage consume a large amount of computational and storage resources.

While knowledge updating with the distillation process can dynamically adjust the student network's learning difficulty,
it can also lead to a decrease in the stability of the knowledge learned by the student. Yun \textit{et al.} have experimentally demonstrated that increasing the intra-class consistency of network prediction could improve distillation performance \cite{yun_regularizing_2020}. Furthermore, this paper also found that allowing the teacher network to change the output information entropy (i.e., changing the stability of the teacher network output) in an adaptive manner resulted in a significant performance decrease. 
Therefore, it is vital to improving the distillation performance by reducing the student network's learning difficulty without changing the knowledge from the teacher network.

In addition, controlling the output entropy of neural networks has been widely used in domain adaptation and semi-supervised learning.
\cite{grandvalet_semi-supervised_2004} improved the performance of semi-supervised learning by minimizing the prediction information entropy of unlabeled data. 
Vu \textit{et al}. first applied entropy minimization to an unsupervised domain adaptive task \cite{vu_advent_2019}. 
However, during the entropy minimization process, the problem of unbalanced gradients in samples with different difficulties can arise. So Chen \textit{et al}. used maximum squares loss to solve this problem \cite{chen_domain_2019}.
Domain adaption is similar to knowledge distillation. It transfers the trained model from the source domain to the target domain \cite{xu_larger_2019}. On the other hand, knowledge distillation migrates knowledge from the teacher network to the student network. Therefore an excellent entropy control algorithm may have a beneficial effect on knowledge distillation.

Inspired by this problem, this paper proposes a knowledge distillation algorithm based on dynamic entropy correction to reduce the performance gap and improve performance, called DynamicKD.
During the distillation, the student network faces two gaps. One is the gap between the student and the teacher outputs, which is usually measured using KL divergence. The other is the gap between the student network outputs and the ground truth labels, and it is traditionally measured using cross-entropy. Both gaps affect the distillation performance and vary with the knowledge distillation process, so this paper calls these losses distillation gaps. 
DynamicKD can reduce these gaps by adjusting the student adaptively, thus reducing the learning difficulty and improving distillation performance.
Firstly, the student network's output entropy is controlled by an entropy controller. This  entropy controller, like the distillation temperature, can increase or decrease the output entropy. Changing the output entropy can adjust the softness and hardness of the output distribution, thus adjusting the distance between the student output distribution and the teacher output distribution and between the student and the ground truth label. Too soft or too hard output distribution could increase the distance. 
This paper proves that distillation gaps have only one local minimum for the adjustable parameter in the entropy controller, and properly adjusting this parameter can reduce distillation gaps.
Then, a knowledge distillation algorithm based on dynamic entropy correction is proposed. It applies an entropy controller to correct the output entropy of the student network, and the entropy controller can be updated using distillation loss in real time. 
Since both cross-entropy loss and KL divergence loss have only one local minimum to the adjustable parameter, the entropy controller can be optimized efficiently by backpropagation during the distillation.
The main contributions of this paper are summarized as the following.

\begin{itemize}
\item Change the distillation gaps with an adjustable parameter. This paper proves, in theory, that these distillation gaps have only one local optimum value for this adjustable parameter and that properly adjusting this parameter can reduce the distillation gaps.

\item A knowledge distillation algorithm based on dynamic entropy correction is proposed. This method corrects the output entropy of the student network and improves the performance by an entropy controller, which can be updated in real-time using distillation losses.

\item Extensive experiments verify the effectiveness and efficiency of the proposed algorithm. The performance comparisons with the state-of-the-art distillation algorithms on CIFAR100 and ImageNet datasets demonstrate the value of the proposed dynamic entropy correction strategy for knowledge distillation.
\end{itemize}

\section{The Proposed Algorithm}
\begin{figure}[h]
\centering
\includegraphics[scale=0.58]{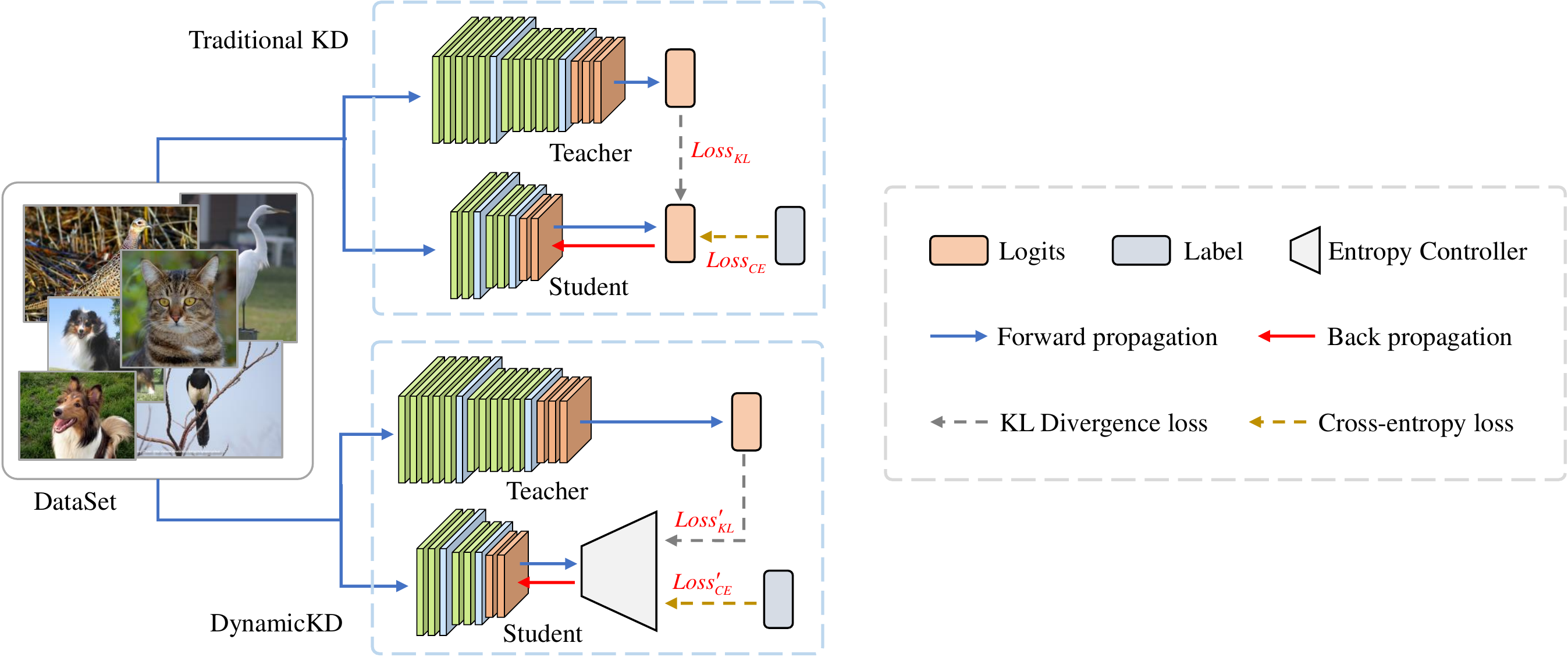}
\caption{Comparison between traditional knowledge distillation algorithm KD and DynamicKD.}
\label{Fig1}
\end{figure}

In knowledge distillation, the distillation gap can affect distillation performance. So, a novel knowledge distillation is proposed. It can reduce the distillation gaps dynamically and improve the distillation performance. 
The main difference between the traditional knowledge distillation algorithm KD and DynamicKD is whether it uses the entropy controller. 
And Figure \ref{Fig1} shows this structure difference.

For the traditional knowledge distillation, the teacher is a trained high-performance network, and the student is a lightweight network to be trained. The student and the teacher generate logits separately. Then the student weights are updated with cross-entropy loss and KL divergence loss. The cross-entropy loss is utilized to measure the gap between the student network output and the ground truth labels. The KL divergence loss is used to measure the output gap between the teacher network and the student network. Unlike the traditional method, DynamicKD applies an entropy controller to correct the output entropy of the student network, 
thus reducing the distillation gap and optimizing the distillation performance.

In the remainder of this section, this paper first analyses the impact of changing the output entropy in the student on the distillation gap in theory. Then, the proposed algorithm DynamicKD and the model reparameterization for the trained student are introduced. Also, this paper compares the proposed algorithm DynamicKD with the traditional knowledge distillation algorithms and describes their differences.

\subsection{Theoretical Analysis about the Effect of Entropy Change on the Distillation Gaps}
During distillation, the output entropy change of neural networks affects the distillation performance \cite{hinton_distilling_2015}, which inspires us to study how the entropy change affects the distillation gaps. To facilitate the analysis, this section begins with some mathematical descriptions. In a classification task with $m$ classes, let the output vector of the neural network be $Z=\{z_1, z_2,\dots , z_m\}$ (this paper does not consider $z_1=z_2=z_3=\dots =z_m$. 
In this case, it is not possible to determine which class the sample belongs to). And the teacher and student network output vectors are denoted as
$Z^{(t)}$ and $Z^{(s)}$, respectively. 
Let the network prediction distribution softened by the distillation temperature $T$ be $P_{T}=\{p_{1,T}, \dots, p_{i,T}, \dots ,p_{m,T}\}$, where $p_{i,T}$ is the \textit{i}-th class's prediction distribution and is defined in Equation (\ref{eq1}).
% The network prediction distribution $P_{i,T}$ of the \textit{i}-th class softened by the distillation temperature $T$ is defined in Equation (\ref{eq1}).
\begin{equation}
    p_{i,T}=\frac{exp(z_i/T)}{\sum_{j=1}^m exp(z_j/T)}
    \label{eq1}
\end{equation}

There are two distillation gaps, which two loss functions can measure. One is the gap between the student and the ground truth labels measured by cross-entropy loss $Loss_{CE}$; the other is the gap between the student and the teacher measured by KL divergence loss $Loss_{KL}$.
Equations (\ref{eq2}) and (\ref{eq3}) define these two losses.
\begin{equation}
    Loss_{CE}=-\sum_{i=1}^{m} y_i log(p_{i,1}^{(s)})
    \label{eq2}
\end{equation}
\begin{equation}
    Loss_{KL}=- T^2 \sum_{j=1}^{m} p_{{j,T}}^{(t)} log(\frac{p_{j,T}^{(s)}}{p_{j,T}^{(t)}})
    \label{eq3}
\end{equation}
where $p_{j,T}^{(s)}$ denotes the softened distribution from the student network on the \textit{i}-th class, $y_i$ represents the ground truth label on the \textit{i}-th class whose value is 0 or 1. A sample only belongs to one class, which means $y_k=1,y_j=0,j \neq k$ when it belongs to the \textit{k}-th class.

Here, a parameter $\alpha$ is applied to change the output entropy of the student network, and the adjusted student network output $z^{(s)^\prime}$ is shown in Equation (\ref{eq4}).
\begin{equation}
    z^{(s)'} = \alpha z^{(s)},\alpha \in (0,+\infty)
    \label{eq4}
\end{equation}

Information entropy is used to measure the uncertainty—the greater the uncertainty, the greater the entropy. The output distribution entropy of a neural network modified by the parameter $\alpha$ is defined below.
\begin{equation}
    H\left(Z, \alpha \right)=-\sum_{j=1}^m p_{j,\alpha,T}^{(s)'} log \left(p_{j,\alpha,T}^{(s)'}\right)
    \label{1-1}
\end{equation}
where $p_{j,\alpha,T}^{(s)'}$ is the softened output distribution of the student network on the \textit{j}-th class modified by the parameter $\alpha$ with the temperature $T$, and it is defined as Equation (\ref{eq6}).
\begin{equation}
    p_{j,\alpha,T}^{(s)'} = \frac{exp(\alpha z_j^{(s)}/T)}{\sum_{l=1}^m exp(\alpha z_l^{(s)}/T)}
    \label{eq6}
\end{equation}

Similar to the distillation temperature $T$, adjusting $\alpha$ can change the certainty of the network prediction results. The higher the parameter $\alpha$, the closer the prediction probability distribution is to the one-hot label and the smaller the output entropy; conversely, the closer the prediction probability distribution is to the uniform distribution and the larger the output entropy. During the network output entropy change, the gap between the student and the teacher and the gap between the student and the ground truth label are also changing. Too large or too small a parameter $\alpha$ will make the output entropy of the network too small or too large, thus increasing these two gaps and increasing the training difficulty of the student.
The relationship between $\alpha$ and KL divergence loss and cross-entropy loss is analyzed in the following.

The KL divergence loss $Loss_{KL}^{'} (\alpha)$ modified by the parameter $\alpha$ is defined as shown in Equation (\ref{eq5}).
\begin{equation}
    Loss_{KL}^{'}(\alpha) = -T^2 \sum_{j=1}^m p_{j,T}^{(t)} log \frac{p_{j,\alpha,T}^{(s)'}}{p_{j,T}^{(t)}}
    \label{eq5}
\end{equation}

The derivative of Equation (\ref{eq5}) with respect to the parameter $\alpha$ is shown in Equation (\ref{eq7}).
\begin{equation}
    \frac{\partial Loss_{KL}^{'}(\alpha)}{\partial \alpha} = -T^2 \sum_{j=1}^m p_{j,T}^{(t)} \frac{p_{j,T}^{(t)}}{p_{j,\alpha, T}^{(s)'}} \frac{\partial p_{j,\alpha,T}^{(s)'}}{\partial \alpha}
    \label{eq7}
\end{equation}

The derivative of Equation (\ref{eq6}) for the parameter $\alpha$ is shown in Equation (\ref{eq8}).
\begin{equation}
\begin{aligned}
\frac{\partial p_{j, \alpha, T}^{(s)^{\prime}}}{\partial \alpha}=
&\frac{1}{T}\frac{\exp \left(\alpha z_{j}^{(s)} / T\right)}{\sum_{l=1}^{m} \exp \left(\alpha z_{l}^{(s)} / T\right)} 
\left(z_{j}^{(s)}-\frac{\sum_{k=1}^{m} z_{k}^{(s)} \exp \left(\alpha z_{k}^{(s)} / T\right)}{\sum_{l=1}^{m} \exp \left(\alpha z_{l}^{(s)} / T\right)}\right) \\
=&\frac{1}{T} p_{j, \alpha, T}^{(s)^{\prime}}\left(z_{j}^{(s)}-\sum_{k=1}^{m} p_{k, \alpha, T}^{(s)^{\prime}} z_{k}^{(s)}\right)
\end{aligned}
\label{eq8}
\end{equation}

Then, from Equation (\ref{eq8}), Equation (\ref{eq7}) can be rewritten as Equation (\ref{eq9}).
\begin{equation}
    \frac{\partial Loss_{KL}^{'} (\alpha)}{\partial \alpha} = 
    T \sum_{j=1}^m 
    (p_{j,T}^{(t)})^2 (\sum_{k=1}^m p_{k,\alpha,T}^{(s)'} z_k^{(s)} - z_j^{(s)})
    \label{eq9}
\end{equation}

Therefore, Equation (\ref{eq9}) is influenced by the weighted average $\overline{f(\alpha)} = \sum_{k=1}^m p_{k,\alpha,T}^{(s)'} z_k^{(s)}$. From Equation (\ref{eq10}), it follows that this weighted average is monotonically increasing.
\begin{equation}
\begin{aligned}
\frac{\partial \overline{f(\alpha)}}{\partial \alpha}=
&\frac{1}{2 T} \sum_{i=1}^{m} \sum_{j=1}^{m}\left(z_{i}^{(s)}-z_{j}^{(s)}\right)^{2} \frac{\exp \left(\alpha z_{i}^{(s)} / T \right) \exp \left(\alpha z_{j}^{(s)} / {T}\right)}{\left(\sum_{l=1}^{m} \exp \left(\alpha z_{l}^{(s)} / {T}\right)\right)^{2}} \\
=&\frac{1}{2T} \sum_{i=1}^m \sum_{j=1}^m (z_i^{(s)} - z_j^{(s)})^2 p_{i,\alpha,T}^{(s)'} p_{j,\alpha,T}^{(s)'} > 0
\end{aligned}
\label{eq10}
\end{equation}

Thus, Equation (\ref{eq9}) is monotonically increasing. When $\alpha \rightarrow 0$, the following can be obtained.
\begin{equation}
\left.p_{k, \alpha, T}^{(s)^{\prime}} z_k^{(s)}\right|_{\alpha \rightarrow 0}=
\left.\frac{\exp \left(\alpha z_k^{(s)} / T\right)}{\sum_{l=1}^m \exp \left(\alpha z_l^{(s)} / T\right)} z_k^{(s)}\right|_{\alpha \rightarrow 0}=
\frac{1}{m} z_k^{(s)}
\label{eq11-1}
\end{equation}

\begin{equation}
\begin{aligned}
\left.\frac{\partial Loss_{KL}{'} (\alpha)}{\partial \alpha}\right|_{\alpha \rightarrow 0} 
= & T \sum_{j=1}^m (p_{j,T}^{(t)})^2 (\left.\sum_{k=1}^m p_{k,\alpha,T}^{(s)'} z_k^{(s)}\right|_{\alpha \rightarrow 0} - z_j^{(s)}) \\
= &T \sum_{j=1}^m (p_{j,T}^{(t)})^2 (\frac{1}{m} \sum_{k=1}^m z_k^{(s)} - z_j^{(s)}) 
= 
% \sum_{k=1}^m p_{k,\alpha,T}^{(s)'} z_k^{(s)}
T \sum_{j=1}^m (p_{j,T}^{(t)})^2 (Z_{avg}^{(s)} - z_j^{(s)})
\end{aligned}
\label{eq11}
\end{equation}

When $\alpha \rightarrow +\infty$, the following can be obtained.
\begin{equation}
\left.p_{k, \alpha, T}^{(s)^{\prime}} z_k^{(s)}\right|_{\alpha \rightarrow+\infty}=\left.\frac{\exp \left(\alpha z_k^{(s)} / T\right)}{\sum_{l=1}^m \exp \left(\alpha z_l^{(s)} / T\right)} z_k^{(s)}\right|_{\alpha \rightarrow+\infty}=\left\{\begin{array}{c} 0 * z_k^{(s)}=0, z_k^{(s)} \neq Z_{\max }^{(s)} \\
1 * z_k^{(s)}=z_k^{(s)}, z_k^{(s)}=Z_{\max }^{(s)}
\end{array}\right.
\label{12-1}
\end{equation}

\begin{equation}
\begin{aligned}
\left.\frac{\partial Loss_{KL}{'} (\alpha)}{\partial \alpha}\right|_{\alpha \rightarrow +\infty} 
= & T \sum_{j=1}^m (p_{j,T}^{(t)})^2 (\left.\sum_{k=1}^m p_{k,\alpha,T}^{(s)'} z_k^{(s)}\right|_{\alpha \rightarrow +\infty} - z_j^{(s)}) \\
= & T \sum_{j=1}^m (p_{j,T}^{(t)})^2 (Z_{max}^{(s)} - z_j^{(s)})
\end{aligned}
\label{eq12}
\end{equation}
where $Z_{avg}^{(s)}$ denotes the average of all elements in the vector $Z^{(s)}$, and $Z_{max}^{(s)}$ represents the maximal element in the vector $Z^{(s)}$. 
Equation (\ref{eq11}) is equivalent to a weighted sum. When the output distributions of the teacher network and the student network are similar, $\frac{\partial Loss_{KL}^{'} (\alpha)}{\partial \alpha}|_{\alpha \rightarrow 0} < 0$. And since $\frac{\partial Loss_{KL}^{'} (\alpha)}{\partial \alpha}|_{\alpha \rightarrow +\infty} > 0$ and $\frac{\partial Loss_{KL}^{'} (\alpha)}{\partial \alpha}$ is monotonically increasing, a solution $\alpha ^{*} \in (0,+\infty)$ can be obtained where Equation (\ref{eq9}) is 0. And because of the monotonic increasing in $\frac{\partial Loss_{KL}^{'} (\alpha)}{\partial \alpha}$, this solution is the only locally optimal solution of $Loss_{KL}^{'}(\alpha)$. So Equation(\ref{eq12-2}) can be got.
\begin{equation}
\frac{\partial {Loss}_{K L}{ }^{\prime}(\alpha)}{\partial \alpha}= 
\begin{cases}\lim _{\Delta \alpha \rightarrow 0} \frac{{Loss}_{K L}{ }^{\prime}(\alpha+\Delta \alpha)-{Loss}_{K L}{ }^{\prime}(\alpha)}{\Delta \alpha}>0 & \alpha>\alpha^* \\ \lim _{\Delta \alpha \rightarrow 0} \frac{{Loss}_{K L}{ }^{\prime}(\alpha+\Delta \alpha)-{Loss}_{K L}{ }^{\prime}(\alpha)}{\Delta \alpha}<0 & \alpha<\alpha^*\end{cases}
\label{eq12-2}
\end{equation}

From Equation (\ref{eq12-2}), Equation (\ref{eq12-3}) can be obtained.
\begin{equation}
\left\{\begin{array}{cc}
{Loss}_{K L}{ }^{\prime}(\alpha+\Delta \alpha)<{Loss}_{K L}{ }^{\prime}(\alpha), \alpha^*<\alpha+\Delta \alpha, \Delta \alpha<0 & \alpha>\alpha^* \\
{Loss}_{K L}{ }^{\prime}(\alpha+\Delta \alpha)<{Loss}_{K L}{ }^{\prime}(\alpha), 0<\alpha+\Delta \alpha<\alpha^*, \Delta \alpha>0 & \alpha<\alpha^*
\end{array}\right.
\label{eq12-3}
\end{equation}

Equation (\ref{eq12-3}) shows that there is always a $\Delta \alpha$ such that ${Loss}_{K L}{ }^{\prime}(\alpha+\Delta \alpha)<{Loss}_{K L}{ }^{\prime}(\alpha)$ holds.

When there is a significant difference in the output distribution of the teacher network and the student network, $\frac{\partial Loss_{KL}^{'} (\alpha)}{\partial \alpha} > 0$ is constant and since $\frac{\partial Loss_{KL}^{'} (\alpha)}{\partial \alpha}$ is monotonically increasing, $Loss_{KL}^{'}(\alpha)$ gets the only locally optimal solution when $\alpha \rightarrow 0$. So Equation (\ref{eq12-4}) can be got.
\begin{equation}
\frac{\partial {Loss}_{K L}{ }^{\prime}(\alpha)}{\partial \alpha}=\lim _{\Delta \alpha \rightarrow 0} \frac{{Loss}_{K L}{ }^{\prime}(\alpha+\Delta \alpha)-{Loss}_{K L}{ }^{\prime}(\alpha)}{\Delta \alpha}>0
\label{eq12-4}
\end{equation}

From Equation (\ref{eq12-4}), Equation (\ref{eq12-5}) can be obtained.
\begin{equation}
{Loss}_{K L}{ }^{\prime}(\alpha+\Delta \alpha)<{Loss}_{K L}{ }^{\prime}(\alpha), \alpha+\Delta \alpha>0, \Delta \alpha<0
\label{eq12-5}
\end{equation}

Equation (\ref{eq12-5}) shows that there is always a $\Delta \alpha$ such that ${Loss}_{K L}{ }^{\prime}(\alpha+\Delta \alpha)<{Loss}_{K L}{ }^{\prime}(\alpha)$ holds.

Therefore, from Equation (\ref{eq12-3}) and (\ref{eq12-5}), 
adjusting $\alpha$ by a suitable $\Delta \alpha$ can reduce the KL divergence loss ${Loss}_{K L}{ }^{\prime}(\alpha)$, i.e., reducing the gap between the student and the teacher.

Similarly, the cross-entropy loss $Loss_{CE}^{'}(\alpha)$ corrected by the parameter $\alpha$ is
\begin{equation}
    Loss_{CE}^{'}(\alpha)=-\sum_{i=1}^m y_i log(p_{i,\alpha,1}^{(s)'})
    \label{eq13}
\end{equation}
 									
Then, the derivative of Equation (\ref{eq13}) for the parameter $\alpha$ can be obtained in Equation (\ref{eq14}).
\begin{equation}
    \frac{\partial Loss_{CE}^{'}(\alpha)}{\partial \alpha} = \sum_{i=1}^m p_{i,\alpha,1}^{(s)'} z_i^{(s)} - z_k^{(s)}
    \label{eq14}
\end{equation}
where $z_k^{(s)}$ denotes the prediction of the student network on the true class $k$. From Equation (\ref{eq10}), $\sum_{i=1}^m p_{i,\alpha,1}^{(s)'} z_i^{(s)}$ is monotonically increasing, so Equation (\ref{eq14}) is monotonically increasing. Equation (\ref{eq14}) values at the two extreme points $\{0,+\infty\}$ can be obtained in Equation (\ref{eq15}).
\begin{equation}
    \frac{\partial Loss_{CE}^{'}(\alpha)}{\partial \alpha}= 
    \left\{
    \begin{array}{cc}
        Z_{avg}^{(s)} - z_k^{(s)} & \alpha \rightarrow 0 \\
        Z_{max}^{(s)} - z_k^{(s)} & \alpha \rightarrow +\infty
    \end{array}
    \right.
    \label{eq15}
\end{equation}

Therefore, when the student network performance is relatively poor, it has a smaller prediction on the true classes $k$, $(\frac{\partial Loss_{CE}^{'}(\alpha)}{\partial \alpha}|_{\alpha \rightarrow 0}=Z_{avg}^{(s)} - z_k^{(s)}>0)$. Because  $\frac{\partial Loss_{CE}^{'}(\alpha)}{\partial \alpha}$ is monotonically increasing, $\frac{\partial Loss_{CE}^{'}(\alpha)}{\partial \alpha}>0$ is constant. Thus, when $\alpha \rightarrow 0$, $Loss_{CE}^{'}(\alpha)$ gets the only locally optimal solution. So Equation (\ref{eq15-2}) can be obtained.
\begin{equation}
\frac{\partial {Loss}_{C E}{ }^{\prime}(\alpha)}{\partial \alpha}=\lim _{\Delta \alpha \rightarrow 0} \frac{{Loss}_{C E}{ }^{\prime}(\alpha+\Delta \alpha)-{Loss}_{C E}{ }^{\prime}(\alpha)}{\Delta \alpha}>0
\label{eq15-2}
\end{equation}

From Equation (\ref{eq15-2}), Equation (\ref{eq15-3}) can be obtained.
\begin{equation}
{Loss}_{C E}{ }^{\prime}(\alpha+\Delta \alpha)<{Loss}_{C E}{ }^{\prime}(\alpha), \alpha+\Delta \alpha>0, \Delta \alpha<0
\label{eq15-3}
\end{equation}

Equation (\ref{eq15-3}) shows that there is always a $\Delta \alpha$ such that ${Loss}_{CE}{ }^{\prime}(\alpha+\Delta \alpha)<{Loss}_{CE}{ }^{\prime}(\alpha)$ holds.

When the student network performance is improved, it has a larger prediction on the true class $k$, $(\frac{\partial Loss_{CE}^{'}(\alpha)}{\partial \alpha}|_{\alpha \rightarrow 0}=Z_{avg}^{(s)} - z_k^{(s)}<0)$. And since $\frac{\partial Loss_{CE}^{'}(\alpha)}{\partial \alpha}|_{\alpha \rightarrow +\infty}>0$ and $\frac{\partial Loss_{CE}^{'}(\alpha)}{\partial \alpha}$ is monotonically increasing, Equation (\ref{eq13}) has the only locally optimal solution $\alpha^{*} \in (0, +\infty)$ where Equation (\ref{eq14}) is 0. So Equation (\ref{eq15-4}) can be got.
\begin{equation}
\frac{\partial {Loss}_{C E}{ }^{\prime}(\alpha)}{\partial \alpha}= \begin{cases}\lim _{\Delta \alpha \rightarrow 0} \frac{{Loss}_{C E}{ }^{\prime}(\alpha+\Delta \alpha)-{Loss}_{C E}{ }^{\prime}(\alpha)}{\Delta \alpha}>0 & \alpha>\alpha^* \\ \lim _{\Delta \alpha \rightarrow 0} \frac{{Loss}_{C E}{ }^{\prime}(\alpha+\Delta \alpha)-{Loss}_{C E}{ }^{\prime}(\alpha)}{\Delta \alpha}<0 & \alpha<\alpha^*\end{cases}
\label{eq15-4}
\end{equation}

From Equation (\ref{eq15-4}), Equation (\ref{eq15-5}) can be obtained.

\begin{equation}
\left\{\begin{array}{cl}
{Loss}_{C E}{ }^{\prime}(\alpha+\Delta \alpha)<{Loss}_{C E}{ }^{\prime}(\alpha), \alpha^*<\alpha+\Delta \alpha, \Delta \alpha<0 & \alpha>\alpha^* \\
{Loss}_{C E}{ }^{\prime}(\alpha+\Delta \alpha)<{Loss}_{C E}{ }^{\prime}(\alpha), 0<\alpha+\Delta \alpha<\alpha^*, \Delta \alpha>0 & \alpha<\alpha^*
\end{array}\right.
\label{eq15-5}
\end{equation}

Equation (\ref{eq15-5}) shows that there is always a $\Delta \alpha$ such that ${Loss}_{C E}{ }^{\prime}(\alpha+\Delta \alpha)<{Loss}_{C E}{ }^{\prime}(\alpha)$ holds.

Therefore, from Equation (\ref{eq15-3}) and (\ref{eq15-5}), 
adjusting $\alpha$ by a suitable $\Delta \alpha$ can reduce the cross-entropy loss ${Loss}_{C E}{ }^{\prime}(\alpha)$, i.e., reducing the gap between the student and the ground truth labels.

In summary, ${Loss}_{KL}{ }^{\prime}(\alpha)$ and ${Loss}_{CE}{ }^{\prime}(\alpha)$ are monotonically increasing, and have an attainable optimal solution. Adjusting the alpha by a suitable alpha can reduce distillation gaps.
For example, when the student performs poorly, reducing the alpha value appropriately (e.g., less than 1) can reduce the distillation gaps.

\subsection{The Knowledge Distillation Based on Dynamic Entropy Correction}
\begin{figure}[h]
\centering
\includegraphics[scale=0.56]{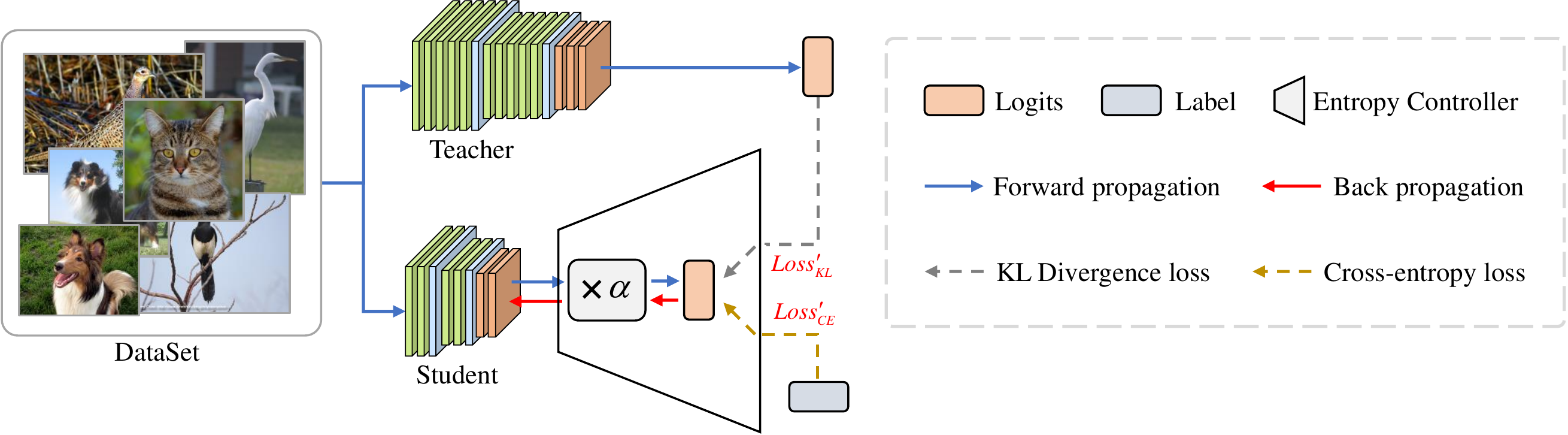}
\caption{Knowledge distillation based on dynamic entropy correction.}
\label{Fig3}
\end{figure}

A proper entropy correction for the student network can improve the knowledge distillation performance. However, the parameter $\alpha$ can be configured arbitrarily within a continuous range at each training epoch. Its configuration can vary as the epoch increases during the training process. So, adjusting the parameter $\alpha$ by hand is impractical in the application process. The configuration space for $\alpha$ grows exponentially with the number of training epochs.
To this end, this paper proposes a knowledge distillation based on dynamic entropy correction. Figure \ref{Fig3} shows the details of the proposed algorithm.

\vspace{0.4cm}

This method can dynamically optimize the entropy correction parameter and compensate for the shortage of manual methods.
Unlike the traditional knowledge distillation algorithms, this algorithm utilizes an entropy controller to correct the output entropy of the student network. 
The output of the entropy controller can be used to calculate KL divergence loss $Loss_{KL}^{'}(\alpha)$ and cross-entropy loss $Loss_{CE}^{'}(\alpha)$ with logits and ground-truth labels. Both losses can update the controller by backpropagation. So, the output entropy of the student network can be corrected in real-time, and this method makes up for the shortage of the manual method.

In addition, $Loss_{KL}^{'} (\alpha)$ and $Loss_{CE}^{'} (\alpha)$ are continuous functions and contain the only locally optimal solutions, which guarantees that the output entropy of the student network can be corrected accurately during the distillation. The distillation error $Loss^{'} (\alpha)$ of the proposed algorithm is defined as the Equation (\ref{eq16}).
\begin{equation}
    Loss^{'}(\alpha) = \beta Loss_{KL}^{'}(\alpha) + Loss_{CE}^{'}(\alpha)
    \label{eq16}
\end{equation}
where $\beta$ is the weight factor for balancing $Loss_{KL}^{'}(\alpha)$ and $Loss_{CE}^{'}(\alpha)$, $Loss_{CE}^{'}(\alpha)$ is the cross-entropy loss with the entropy controller, and $Loss_{KL}^{'}(\alpha)$ is the KL divergence loss with the entropy controller. During the distillation, the entropy controller is dynamically updated by backpropagation of the loss $Loss^\prime (\alpha)$.

\begin{figure}[h]
\centering
\includegraphics[scale=0.6]{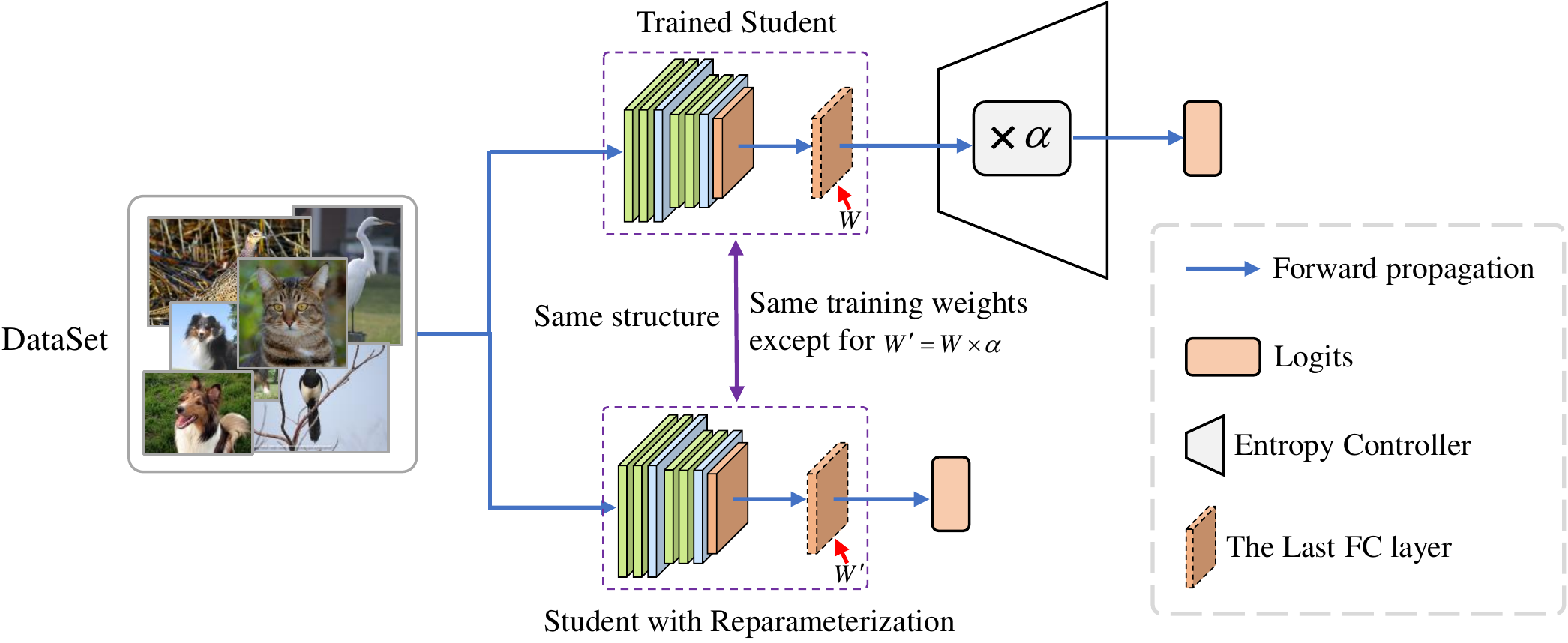}
\caption{Model reparameterization of the student network.}
\label{Fig4add}
\end{figure}

The entropy controller enables the DynamicKD to flexibly adjust the learning difficulty of the student network and optimize the distillation process. 
However, this approach also changes the structure of the student network, and taking the trained student network into the application may produce wrong class similarity prediction.
For this reason, this paper performs model reparameterization on the last fully connected layer of the trained student network to recover its performance. This approach is shown in Figure \ref{Fig4add}. 

\vspace{0.4cm}

As shown in Figure \ref{Fig4add}, the method recomputes the weight $W$ of the last fully connected layer. The weight $W$ has been trained with the proposed distillation method DynamicKD. 
It obtains the reparameterization weight $W^\prime$ by the parameter $\alpha$ in the entropy controller and the weight $W$.
Because there is no nonlinear transformation between the last fully connected layer and the entropy controller, the output of the student network is the same as the one with the entropy controller. 
The reparameterization merges alpha into the final layer of the network; however, this does not mean that the improvement comes from simply changing the weight of the student.
Firstly, the changing and adjusting of the entropy controller are very complex; it makes dynamic adjustments in response to dynamic changes in the distillation gaps and does not simply change the student weights.
Secondly, Equations (\ref{eq12-3}), (\ref{eq12-5}), (\ref{eq15-3}), and (\ref{eq15-5}) also show that a suitable change in alpha can reduce the distillation gaps.

The complete framework of the proposed algorithm is shown in Algorithm \ref{algorithm1}. It first adds the entropy controller to the student network to be trained. Then with the help of the trained teacher network, the knowledge distillation on the student network is done. Finally, network parameterization removes the entropy controller from the student network, and the trained student network is returned.

\begin{algorithm}
	\small
	\caption{The framework for DynamicKD}
	\label{algorithm1}
	\KwIn{Data set $DataSet$, trained teacher network $N_{teacher}$, untrained student network $N_{student}$, number of training iterations $epochs$.}
	\KwOut{The trained student network $N_{student}^{'}$.}
    $epoch=0$\;
    Add the entropy controller to the $N_{student}$ by Equation (\ref{eq4})\;
    \While{$epoch < epochs$}{
        Forward propagation of $N_{teacher}$ and $N_{student}$ generating soft labels\;
        $Loss^{'}_{KL}(\alpha), Loss^{'}_{CE}(\alpha) \leftarrow$ Calculate KL divergence loss and cross-entropy loss by Equation (\ref{eq5}) and (\ref{eq13})\;
        % $Loss^{'}_{CE}(\alpha) \leftarrow$ Calculate the cross-entropy loss by Equation (\ref{eq13})\;
        $Loss^{'}(\alpha) \leftarrow$ Calculate the total loss by Equation (\ref{eq16})\;
        Loss $Loss^{'}(\alpha)$ Backpropagation\; 
        $epoch = epoch + 1$\;
    }
    Remove the entropy controller from the student network by network reparameterization\;
    Return the trained student network $N_{student}^{'}$.
\end{algorithm}

\vspace{-0.4cm}
Finally, although the entropy controller, like the distillation temperature, can adjust the information entropy of the student network, it is different from the simple temperature adjusting method. DynamicKD can reduce the student's training difficulty and improve the distillation performance by dynamically adjusting the distillation gaps during the distillation process. In contrast, the distillation temperature is static during the process. So DyanmicKD performs better than the traditional distillation algorithm KD. In addition, the distillation algorithm using learnable temperature for the student performs worse than DyanmicKD. Because the entropy controller on DynamicKD affects both the KL divergence loss and cross-entropy loss, while the distillation temperature on KD only affects the KL divergence loss. The synchronous adjustment of both losses can avoid the excessive reduction of the KL divergence loss or the cross-entropy loss. The experiment results show that the excessive reduction is harmful to DynamicKD.

\section{The Experimental Setup and Results Analysis}
In this section, this paper first presents the experiments' parameter settings. Then the experiments on the CIFAR100 \cite{krizhevsky_learning_2009} and ImageNet \cite{deng_imagenet_2009} benchmark datasets are conducted to test the proposed algorithm comprehensively. Next, self-distillation experiments are performed to investigate the generalization performance of the proposed algorithm DynamicKD on other knowledge distillation tasks. After that, the cross-validation experiments on the CIFAR100 are conducted to test the effectiveness of the proposed algorithm. Finally, a series of performance analysis experiments are conducted to understand the proposed algorithm further.

\subsection{The Datasets and the Compared Algorithms}
The benchmark datasets used in this paper are CIFAR100 \cite{krizhevsky_learning_2009} and ImageNet \cite{deng_imagenet_2009} datasets. CIFAR100 contains 100 classes of color images, and each one consists of 500 training samples and 100 test samples. ImageNet has 1,000 classes of color images and consists of 1.28 million training samples and 50,000 validation samples. 

To evaluate the proposed algorithm, this paper chooses a variety of state-of-the-art knowledge distillation algorithms FitNet \cite{romero_fitnets_2015}, AT \cite{komodakis_paying_2017}, VID \cite{ahn_variational_2019}, RKD \cite{park_relational_2019}, PKT \cite{passalis_learning_2018}, CRD \cite{tian_contrastive_2020}, WSLD \cite{zhou_rethinking_2021}, NST \cite{huang_like_2019}, FSP \cite{yim_gift_2017}, Overhaul \cite{heo_comprehensive_2019}, SP \cite{tung_similarity-preserving_2019}, CC \cite{peng_correlation_2019}, AB \cite{heo_knowledge_2019}, FT \cite{kim_paraphrasing_2018}, ReviewKD \cite{chen_distilling_2021}, GLD \cite{kim_distilling_2021}, `CS KD' \cite{yun_regularizing_2020} and `PS KD' \cite{kim_self-knowledge_2021} and advanced label regularization algorithms LS \cite{szegedy_rethinking_2016}, Soft Boot \cite{reed_training_2015}, Hard Boot \cite{reed_training_2015}, Disturb Label \cite{xie_disturblabel_2016}, and OLS \cite{zhang_delving_2021}.

\subsection{The Parameters Settings}
In this paper, the parameter settings for the CIFAR100 are the same as Tian \cite{tian_contrastive_2020}. 
The optimizer is SGD with an initial learning rate is 0.05, a momentum of 0.9, and a weight decay of 0.0005. The learning rate is multiplied by 0.1 on the 150th, 180th, and 210th epochs. The batch size and epochs are 64 and 240.
The settings for ImageNet are the same as Heo  \cite{heo_comprehensive_2019}. 
The optimizer is SGD with an initial learning rate of 0.1, a momentum of 0.9, and a weight decay of 0.0001. The learning rate is multiplied by 0.1 on the 30th, 60th, and 90th epochs. The batch size and epochs are 128 and 100.
The entropy controller's initial value of the learnable parameter alpha is 1. It may be better to set different initial values for different distillation experiments. However, determining an appropriate initial value requires many experiments, making the algorithm's application more difficult. So, this paper sets the initial value to 1, which does not scale the student's output.
In addition, the proposed algorithm DynamicKD has two hyperparameters. They are the weighting factor $\beta$ and the distillation temperature $T$. And they can affect the distillation performance. The following two subsections will analyze their effect on the proposed algorithm.

\subsubsection{The Effect of Distillation Temperature $T$ on DynamicKD}
\begin{figure}[h]
\centering
\subfloat[The result on resnet32x4 $\rightarrow$ vgg8] {\includegraphics[scale=0.115]{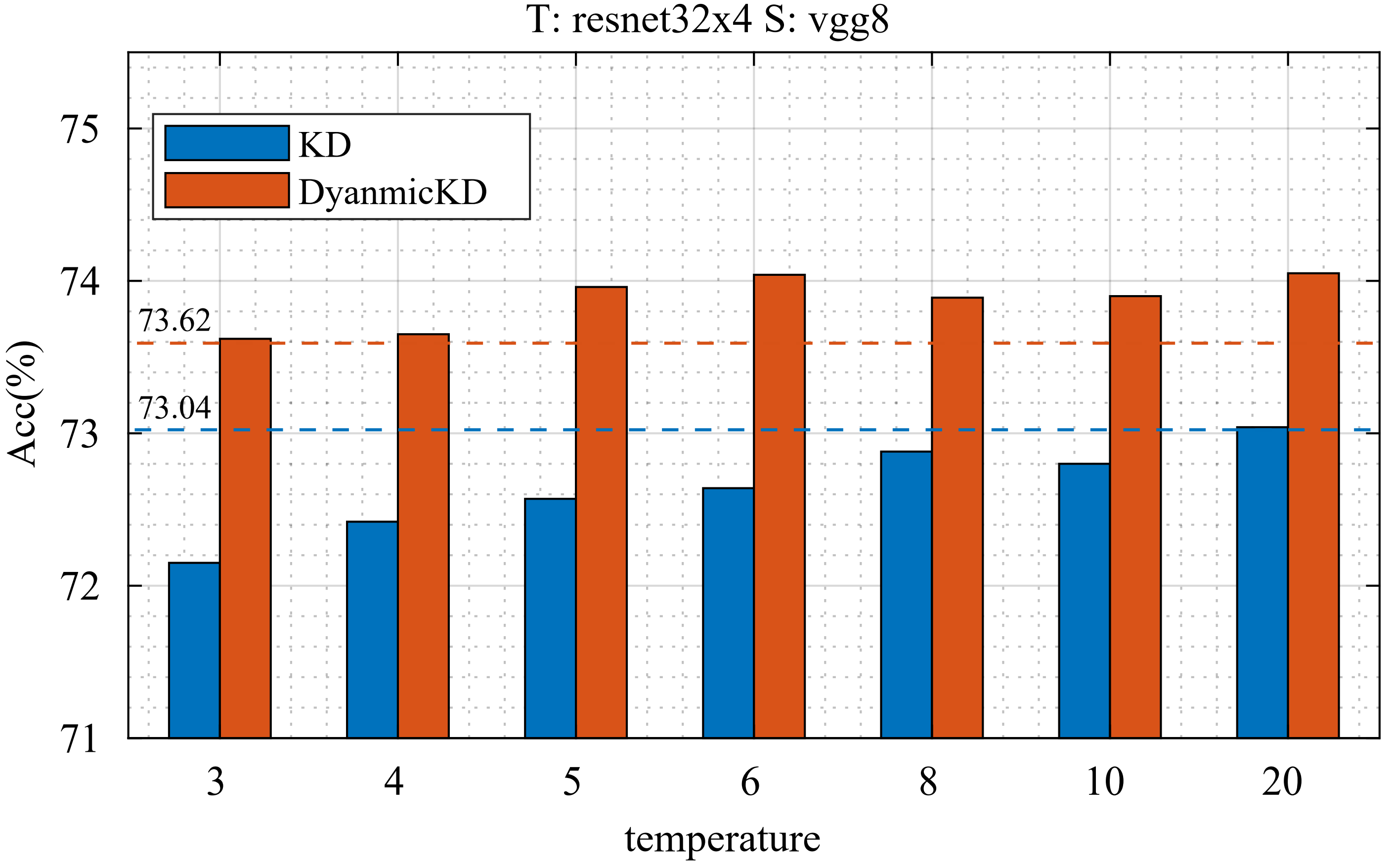}}
\subfloat[The result on vgg13 $\rightarrow$ vgg8] {\includegraphics[scale=0.115]{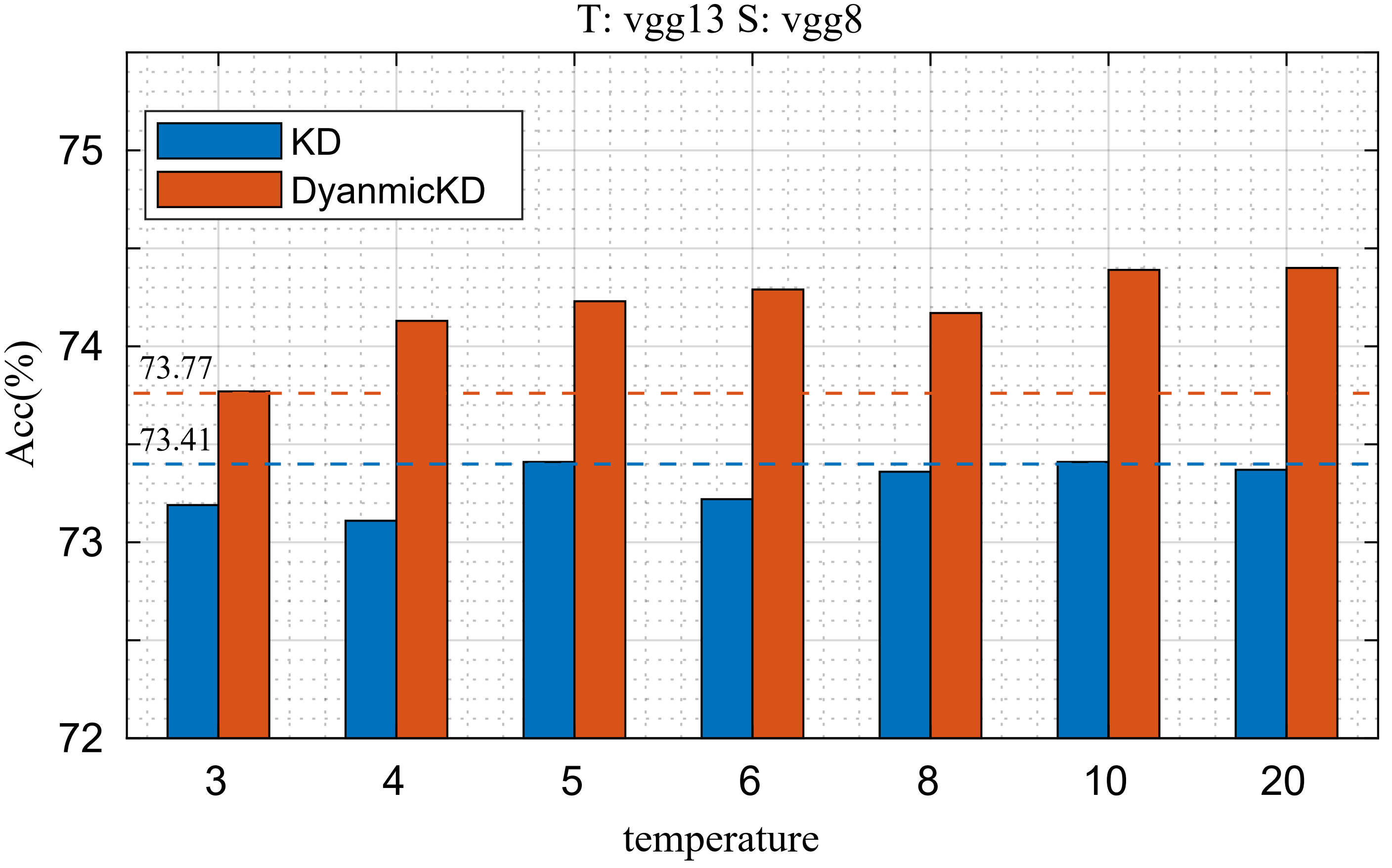}}
\caption{Performance variation curves of DynamicKD and KD at different distillation temperatures.}
\label{Fig10}
\end{figure}

The proposed algorithm DynamicKD can adaptively adjust the output entropy of the student network. At the same time, traditional knowledge distillation algorithms can also change the output entropy of the networks with the distillation temperature. So, to investigate the effects of different distillation temperatures on DynamicKD, this paper tests the performance of DynamicKD and traditional knowledge distillation KD at various distillation temperatures. The experiments are conducted on the CIFAR100 dataset at the distillation experiments resnet32x4 $\rightarrow$ vgg8 and vgg13 $\rightarrow$ vgg8. The experimental results are shown in Figure \ref{Fig10}.
\vspace{-0.2cm}
\vspace{-0.2cm}

Figure \ref{Fig10} shows that DynamicKD outperforms KD at all distillation temperatures. Even the optimal performance achieved by KD does not reach the lowest one obtained by DynamicKD at all distillation temperatures. In the distillation experiment resnet32x4 $\rightarrow$ vgg8, the lowest performance from the proposed algorithm (73.62\%) is 0.58 points higher than the optimal performance from KD (73.04\%) at all temperatures. In the distillation experiment vgg13 $\rightarrow$ vgg8, the proposed algorithm outperforms KD by 0.36 points. This result demonstrates the effectiveness and efficiency of the proposed algorithm.

Although 
KD and 
DynamicKD can improve the distillation performance by adjusting the output entropy of the network, the proposed algorithm is different from the simple distillation temperature adjustment method. 
There is a significant performance gap between 
KD
and DynamicKD. 
In addition, knowledge distillation experiments usually use typical distillation temperatures. For a fair comparison with other algorithms, the optimal distillation temperature is not used in this paper, but the typical distillation temperature is used ($T=4$ in CIFAR100, $T=2$ in ImageNet) \cite{zhou_rethinking_2021}.

\subsubsection{The Effect of Weight Factor $\beta$ on DynamicKD}
This section conducts the parameter analyses for the weighting factor $\beta$.
Firstly, this paper shows the experiments on ImageNet. The teacher network is resnet34, and the student network is resnet18. The experimental results are shown in Table \ref{table1}.
\begin{table}[h]
\footnotesize
\centering
\caption{
Analysis experiments of DynamicKD on ImageNet for the parameter $\beta$.
}
\vspace{-0.1cm}
\setlength{\tabcolsep}{4mm}{
\renewcommand{\arraystretch}{1}
\begin{tabular}{|c|c|c|c|c|c|c|}
\hline
$\beta$ & 0.50      & 0.75      & 1.00      & 1.25     & 1.50      & 1.75    \\ \hline
top-1   & 72.108    & 71.832    & 72.194    & 72.228   & \textbf{72.548}    & 72.420  \\
top-5   & 90.568    & 90.746    & 90.730    & 90.836   & \textbf{90.860}    & 90.856  \\ \hline
\end{tabular}}
\label{table1}
\end{table}
\vspace{-0.2cm}

As can be seen from the table, DynamicKD obtains the best classification accuracy (72.548\% and 90.860\%) on top-1 and top-5 when $\beta=1.50$, and obtains the worst ones (72.108\% and 90.568\%) on top-1 and top-5 when $\beta=0.50$. This indicates that the choice of the $\beta$ parameter affects the DynamicKD, and an appropriate value is beneficial to the proposed algorithm, while an inappropriate one is harmful.

In addition, this paper also analyzes the effect of $\beta$ on the CIFAR100, where the teachers are resnet32x4 and resnet56, the students are resnet8x4 and vgg8. The experimental results are shown in Table \ref{table2}.

\begin{table}[h]
\footnotesize
\centering
\caption{
Analysis experiments of DynamicKD on CIFAR100 training set for the parameter $\beta$.
}
\vspace{-0.1cm}
\setlength{\tabcolsep}{5mm}{
\renewcommand{\arraystretch}{1}
\begin{tabular}{|c|c|c|c|c|c|}
\hline
$\beta$                             & 0.50  & 0.75  & 1.00           & 1.25  & 1.50   \\ \hline
resnet32x4 $\rightarrow$ resnet8x4  & 75.75 & 75.82 & \textbf{76.06} & 75.90 & 75.97  \\
resnet56 $\rightarrow$ vgg8         & 74.19 & 74.30 & \textbf{74.74} & 74.55 & 74.17  \\ \hline
\end{tabular}}
\label{table2}
\end{table}
\vspace{-0.2cm}

DynamicKD achieved the best performance when $\beta=1.00$, 
and its performance drops when $\beta$ decreases or increases.
Again, this shows the importance of choosing the appropriate weighting factor for DynamicKD. In the following experiments, this paper set $\beta=1.00$ for CIFAR100 and 1.50 for ImageNet.

\subsection{The Experiment results and analysis on the Benchmark Datasets}
This paper conducts extensive experiments on the medium-scale benchmark dataset CIFAR100 and the large-scale benchmark dataset ImageNet. And this paper compares the experimental results with various advanced distillation algorithms to comprehensively understand the proposed algorithm’s performance.
Table \ref{table3} shows the classification accuracies of DynamicKD and the peer algorithms on the CIFAR100. The accuracy is followed by the standard deviation. 

\begin{table}[h]
\footnotesize
\centering
\caption{Top-1 classification accuracies (\%) on the CIFAR100.}
\vspace{-0.1cm}
\setlength{\tabcolsep}{1.5mm}{
\renewcommand{\arraystretch}{1.}
\begin{tabular}{|c|c|c|c|c|c|c|c|}
\hline
Method & \multicolumn{4}{c|}{Same   architecture}   & \multicolumn{3}{c|}{Different   architecture} \\ \cline{1-8} 
teacher   & vgg13 & resnet32x4 & resnet56 & resnet110 & resnet56 & vgg13    & vgg13     \\
student   & vgg8  & resnet8x4  & resnet20 & resnet20  & vgg8     & resnet20 & resnet8x4 \\ \hline
teacher   & 75.13 & 79.24      & 72.29    & 74.01     & 72.29    & 75.13    & 75.13     \\
student   & 70.66 & 72.58      & 69.54    & 69.54     & 70.66    & 69.54    & 72.58     \\ \hline
KD        & 73.44(0.12) & 73.42(0.24)      & 70.79(0.21)    & 70.65(0.28)     & 73.43(0.22)    & 69.40(0.12)    & 73.74(0.39)     \\
FitNet    & 71.36(0.25) & 73.32(0.15)      & 69.11(0.33)    & 68.74(0.28)     & 70.03(0.36)    & 68.77(0.34)    & 72.76(0.17)     \\
AT        & 71.33(0.14) & 73.06(0.23)      & 70.33(0.32)    & 70.37(0.11)     & 70.69(0.16)    & 66.00(0.31)    & 71.06(0.14)     \\
VID       & 71.46(0.29) & 73.23(0.22)      & 69.99(0.19)    & 70.25(0.28)     & 72.44(0.16)    & 69.30(0.24)    & 72.40(0.16)     \\
RKD       & 71.18(0.29) & 72.09(0.25)      & 69.52(0.15)    & 69.45(0.19)     & 71.91(0.23)    & 68.53(0.20)    & 72.12(0.22)     \\
PKT       & 73.01(0.28) & 73.79(0.26)      & 70.42(0.21)    & 70.40(0.19)     & 72.60(0.09)    & 70.19(0.25)    & 73.65(0.29)     \\
SP        & 72.67(0.18) & 72.82(0.26)      & 70.43(0.21)    & 70.16(0.20)     & 73.37(0.22)    & 68.47(0.20)    & 73.10(0.25)     \\
CC        & 70.68(0.26) & 72.43(0.31)      & 69.22(0.12)    & 69.03(0.17)     & 70.65(0.24)    & 69.37(0.13)    & 72.57(0.24)     \\
AB        & 70.94(0.25) & 72.70(0.24)      & 69.38(0.14)    & 69.62(0.20)     & n/a      & n/a      & n/a       \\
FT        & 70.58(0.21) & 73.00(0.22)      & 70.00(0.24)    & 69.73(0.21)     & 70.47(0.26)    & 67.34(0.11)    & 72.29(0.13)     \\
NST       & 71.43(0.15) & 73.42(0.31)      & 69.54(0.21)    & 69.48(0.24)     & 68.85(0.22)    & 60.73(0.27)    & 69.34(0.23)     \\
CRD       & 73.66(0.10) & 75.19(0.32)      & 71.17(0.13)    & 71.25(0.14)     & 74.19(0.17)    & 70.53(0.23)    & 74.85(0.08)     \\
WLSD      & 73.48(0.12) & 75.07(0.12)      & 71.59(0.20)    & 71.54(0.22)     & 74.12(0.13)    & 70.42(0.23)    & 75.30(0.24)     \\ \hline
DynamicKD & \textbf{74.13(0.07)} & \textbf{76.06(0.20)}      & \textbf{71.82(0.23)}    & \textbf{71.71(0.12)}    
& \textbf{74.74(0.21)}    & \textbf{70.81(0.23)}    & \textbf{76.01(0.28)}     \\ \hline
\end{tabular}}
\label{table3}
\end{table}
\vspace{-0.2cm}

The experiments consist of different teacher-student pairs with VGG-like networks \cite{simonyan_very_2014} and ResNet-like networks \cite{he_deep_2016}. These experiments are divided into two categories according to whether the teacher and student networks have the same structure. In Table \ref{table3}, the left column compares the same network experiments, and the right column contrasts the different ones.
Table \ref{table3} contains seven knowledge distillation experiments.
Four of them use different teacher and student network structures. The remaining three use different teacher and student network structures. 
The first three rows of the table show the network structure information. The fourth and fifth rows show the prediction accuracy of the teacher and student networks. 
The optimal experimental results are indicated using the bolded font in each distillation experiment.

In the distillation experiments with the same structure,
CRD performed well in the vgg13 $\rightarrow$ vgg8 experiment, achieving 73.66\%. 
This method trains students to obtain more information from the teacher through contrastive learning, and richer information enables students to learn better. 
However, the proposed algorithm improves 0.47 points over the most advanced distillation algorithm CRD and 0.69 points over the traditional method KD at the distillation vgg13 $\rightarrow$ vgg8. 
In the experiments with different structures, 
WLSD performed well in the vgg13 $\rightarrow$ resnet8x4 experiments, achieving 75.30\%. It analyses knowledge distillation from bias-variance tradeoff and balances bias and variance by the adaptive weighting regularization samples. This novel idea brings performance improvement. However,
the algorithm improves WLSD by 0.71 points and the traditional distillation method KD by 2.27 points at the distillation vgg13 $\rightarrow$ resnet8x4. 
In addition, the standard deviation of DynamicKD is similar to other algorithms.
These results demonstrate the effectiveness and efficiency of the proposed algorithm on the medium-sized dataset.

Table \ref{table4} shows the knowledge distillation results of the proposed algorithm and the state-of-the-art algorithms on the large-scale benchmark dataset ImageNet. This table compares top-1 and top-5 classification accuracies. In this case, the teacher network is resnet34, and the student network is resnet18. The experimental results of the peer algorithms are from Zhou \cite{zhou_rethinking_2021}, and this paper uses the same trained teacher network supported by the Pytorch library \cite{paszke_pytorch_2019} as the peer algorithms.
\begin{table}[h]
\footnotesize
\centering
\caption{Top-1 and Top-5 classification accuracies (\%) on the ImageNet. 
}
\vspace{-0.1cm}
\setlength{\tabcolsep}{5.mm}{
\renewcommand{\arraystretch}{1.}
\begin{tabular}{|c|c|c|}
\hline
Method    & Top-1   Accuracy & Top-5   Accuracy \\ \hline
Teacher (resnet34)   & 73.31       & 91.42       \\
Student (resnet18)   & 69.75       & 89.07       \\ \hline
KD        & 70.67       & 90.04       \\
AT        & 71.03       & 90.04       \\
NST       & 70.29       & 89.53       \\
FSP       & 70.58       & 89.61       \\
RKD       & 70.40       & 89.78       \\
Overhaul  & 71.03       & 90.15       \\
CRD       & 71.17       & 90.13       \\
ReviesKD  & 71.61       & 90.51       \\
GLD       & 71.63       & 70.53       \\
WLSD      & 72.04       & 90.70       \\ \hline
DynamicKD & \textbf{72.55}       & \textbf{90.86}       \\ \hline
\end{tabular}}
\label{table4}
\end{table}
\vspace{-0.2cm}

The proposed algorithm surpasses all the comparison algorithms in top-1 and top-5 accuracy. It improves 0.51 points in top-1 accuracy over the state-of-the-art algorithm WLSD, proving the effectiveness and efficiency of the proposed method on the large-scale benchmark dataset ImageNet.

\begin{figure}[h] 
\centering
\subfloat[Variation curves of KD at different training stages]{\includegraphics[scale=0.102]{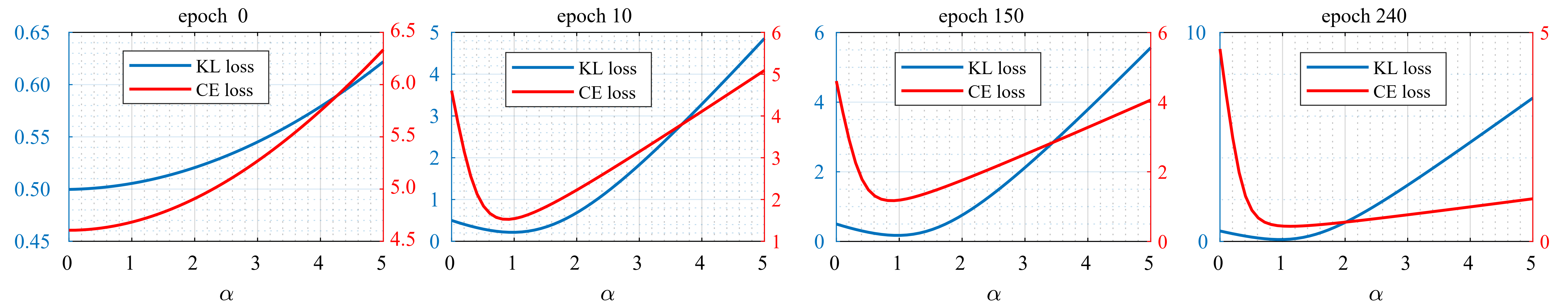}} \hfill
\subfloat[Variation curves of DynamicKD at different training stages]{\includegraphics[scale=0.102]{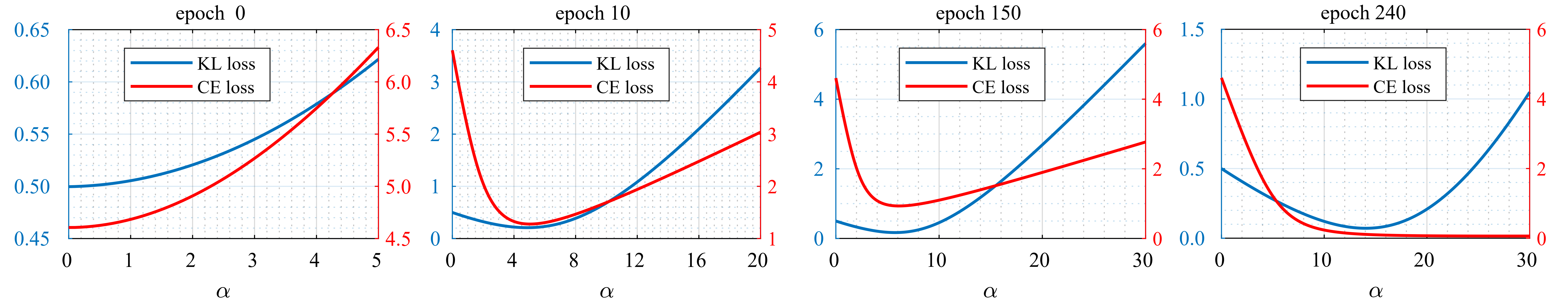}}
\caption{Variation curves of distillation gaps with the parameter $\alpha$ at different training stages.}
\label{Fig5}
\end{figure}

DynamicKD can affect the distillation gaps with the help of the parameter $\alpha$. To understand the effect of DynamicKD on the distillation gaps, this paper compares the variation curves of distillation gaps with the parameter $\alpha$. The result is shown in Figure \ref{Fig5}, where the distillation gaps measured by the KL divergence loss and the cross-entropy loss.

\vspace{0.2cm}
\vspace{-0.2cm}

As shown in Figure \ref{Fig5}, the experiments compare the traditional knowledge distillation algorithm KD and the DynamicKD algorithm at different training stages. This experiment chooses four training stages, epoch 0, epoch 10, epoch 150, and epoch 240. The KL divergence loss (KL) and cross-entropy loss (CE) profiles with the entropy control parameters are depicted at each epoch. The teacher network is resnet32x4; the student network is resnet8x4. DynamicKD's gaps curves are different from KD's. As the training proceeds, the minimum values of the two gaps in KD remain around $\alpha = 1$, while the minimum values in DyanmicKD keep increasing. Even at epoch 240, the minimum value of the cross-entropy loss in Dynamic KD converges to 0 nearly. This phenomenon indicates that DynamicKD with the entropy controller achieves better convergence.

\begin{figure}[h]
\centering
\subfloat[KL Divergence]{\includegraphics[scale=0.77]{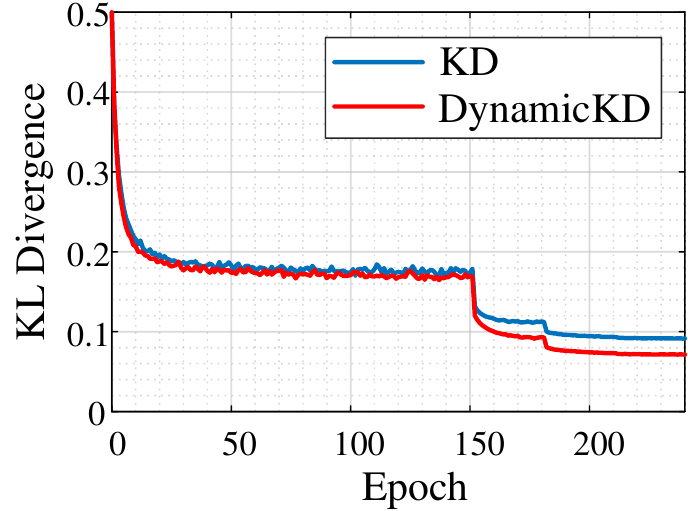}} \qquad
\subfloat[cross entropy]{\includegraphics[scale=0.77]{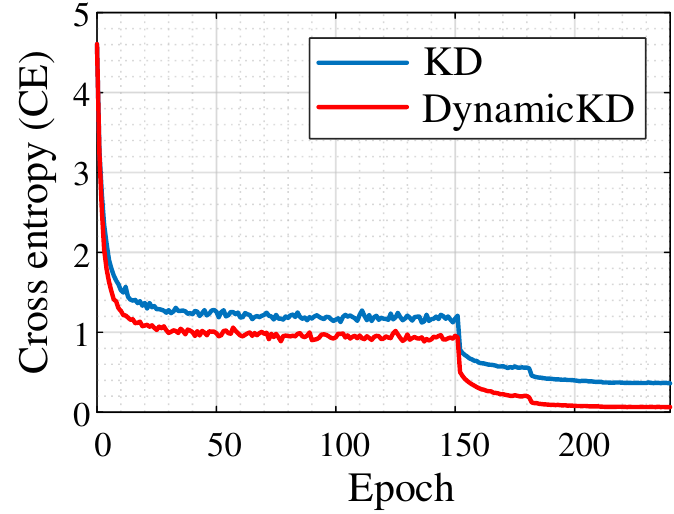}}
\caption{Variation curves of two distillation gap minimums for KD and DynamicKD during the distillation.}
\label{Fig7}
\end{figure}

The entropy controller used in DynamicKD allows the student network to adjust distillation gaps according to different training stages. Figure \ref{Fig7} shows the variation curves of the minimum values of the two distillation gaps in Figure \ref{Fig5} during the training process. 
Compared to the traditional knowledge distillation algorithm KD, DynamicKD can reduce the distillation gaps.
\vspace{-0.2cm}

The reduction in KL divergence loss indicates that the student network can learn better from what the teacher network has learned. The decrease in cross-entropy loss suggests that the student network can better predict the samples.
In addition, the result from the figure shows that this method is different from simply reducing the training loss, which can lead to a severe gradient disappearance problem. In contrast, the proposed method indirectly reduces the KL divergence loss and CE loss of knowledge distillation by adjusting the output entropy of the student network. It does not affect the teacher's output, and
the knowledge learned by the teacher is hardly grasped entirely by the student network (KL divergence loss tends to be 0). Therefore, the proposed algorithm does not lead to a severe gradient disappearance problem due to the unrestricted reduction of the training loss.

\subsection{The Experiment results and analysis on the Self-distillation Tasks}
Self-knowledge distillation is a particular knowledge distillation in which the teacher and student networks have the same network structure. This algorithm is proposed to improve the performance of the student network when the resources are limited, or the high-performance teacher network is unavailable. Firstly, it trains the student network to improve their performance. Then, it distillates the knowledge from the trained network to another student network with the same network structure. In addition, this paper also compares DynamicKD with the label regularization algorithm. Label regularization and knowledge distillation use a similar approach to neural network training, and they both use inter-class similarity information to improve the network’s performance. In self-distillation, the similarity information is from the neural network trained without knowledge distillation (w/o KD). In label regularization, the similarity information comes from uniform distribution or the network output of the previous epoch.

This paper conducts experiments on the CIFAR100 dataset. The performance of the proposed algorithm is compared with the advanced knowledge distillation algorithms (CS KD \cite{yun_regularizing_2020} and PS KD \cite{kim_self-knowledge_2021}) and advanced label regularization algorithms (LS \cite{szegedy_rethinking_2016}, Soft Boot \cite{reed_training_2015}, Hard Boot \cite{reed_training_2015}, Disturb Label \cite{xie_disturblabel_2016}, and OLS \cite{zhang_delving_2021}). It compares the prediction accuracy (Acc) and the performance improvement (Gain) compared to the w/o KD on the three neural networks (resnet20, resnet8x4, and vgg8). The results are shown in Table \ref{table5}, where the distillation parameters used are the same as those in Table \ref{table3}, and all classification accuracies are the average of five runs. The accuracy is followed by the standard deviation.
% \vspace{0.2cm}
\begin{table}[ht]
\footnotesize
\centering
\caption{
Classification accuracies (\%) with self-knowledge distillation and label regularization on the CIFAR100.
}
\vspace{-0.1cm}
\setlength{\tabcolsep}{4mm}{
\renewcommand{\arraystretch}{1}
\begin{tabular}{|c|c|c|c|c|c|c|c|}
\hline
\multicolumn{2}{|c|}{\multirow{2}{*}{Method}} & \multicolumn{2}{c|}{vgg8} & \multicolumn{2}{|c|}{resnet20} & \multicolumn{2}{c|}{resnet8x4} \\ \cline{3-8} 
\multicolumn{2}{|c|}{}                      & Acc          & Gain    & Acc          & Gain   & Acc          & Gain          \\ \hline
None                    & w/o KD            & 70.78        & 0.00    & 69.49        & 0.00   & 72.60        & 0.00          \\ \hline
\multirow{5}{*}{LR}     & LS                & 70.30(0.50)  & -0.48   & 69.37(0.09)  & -0.12  & 72.87(0.35)  & 0.27          \\
                        & Soft Boot         & 71.02(0.39)  & 0.24    & 69.18(0.16)  & -0.31  & 72.35(0.23)  & -0.25         \\
                        & Hard Boot         & 70.21(0.17)  & -0.57   & 69.07(0.08)  & -0.42  & 72.23(0.04)  & -0.37         \\
                        & Disturb Label     & 70.30(0.31)  & -0.48   & 69.32(0.44)  & -0.17  & 72.38(0.25)  & -0.22         \\
                        & OLS               & 70.38(0.25)  & -0.40   & 69.10(0.12)  & -0.39  & 73.12(0.17)  & 0.52          \\ \hline
\multirow{4}{*}{Self-KD}& KD                & 71.31(0.10)  & 0.53    & 69.22(0.14)  & -0.27  & 72.67(0.08)  & 0.07          \\
                        & CS KD             & 71.11(0.26)  & 0.33    & 68.09(0.42)  & -1.40  & 73.48(0.31)  & 0.88          \\
                        & PS KD             & 72.46(0.15)  & 1.68    & 69.98(0.29)  & 0.49   & 73.31(0.11)  & 0.71          \\
                        & DynamicKD         & \textbf{72.70(0.11)} 
                                            & \textbf{1.92}          & \textbf{70.81(0.16)} 
                                            & \textbf{1.32}                                  & \textbf{73.63(0.13)} & \textbf{1.03} \\ \hline
\end{tabular}}
\label{table5}
\end{table}
\vspace{-0.2cm}

The experiments show that although label regularization methods are poor overall, they perform better than the self-distillation methods under specific conditions. For example, the excellent label regularization method OLS outperforms the traditional knowledge distillation method KD on the resnet8x4 network. 
OLS can dynamically change soft labels based on the model's predictions. Such soft labels can carry richer knowledge of inter-category similarities than traditional label smoothing, thus improving performance. However, the smoothed label for each category may not reflect the differences between samples well. So OLS performs poorly in some experiments, such as vgg8 and resnet20.
The state-of-the-art self-distillation algorithm `PS KD' show excellent performance on the vgg8 and resnet20 networks. 
It achieves a performance gain of 1.68 points on the vgg8 network, and 0.49 points on the resnet20 network.
`PS KD' uses a progressive approach combining ground truth labels and past predictions to produce soft targets, which are rich in information to facilitate the student's training.
However, the proposed algorithm outperforms these excellent algorithms on these three networks. It has a gain of nearly 2 points on the vgg8 network. Even on the challenge resnet20 network, the proposed algorithm improves by 1.32 points. 
In addition, the standard deviation of the proposed method is similar to other algorithms.
These experiments demonstrate the effectiveness and efficiency of the proposed algorithm.

\subsection{The Cross-validation Experiments on the CIFAR100}
In this section, the performance of the proposed algorithm with 6-fold cross-validation experiments is analyzed. Cross-validation experiment can reduce the impact of the dataset's division on algorithm performance evaluation and facilitates an objective evaluation of the algorithm's performance. However, since the CIFAR100 dataset used in this paper is usually divided by a predetermined division method (a training set with 50,000 samples and a testing set with 10,000 samples), this paper combines the training set and test set into a dataset containing 60,000 samples and performs 6-fold cross-validation experiments on this dataset. The teacher networks are resnet32x4 and resnet56, and the student networks are resnet8x4 and vgg8. This paper compares the proposed algorithm DynamicKD with the traditional knowledge distillation algorithm KD and the state-of-the-art knowledge distillation algorithm CRD. The classification accuracies are shown in Table \ref{table6}. The accuracy is followed by the standard deviation.

% \vspace{0.2cm}
\begin{table}[ht]
\footnotesize
\centering
\caption{
Cross-validation experiments on the CIFAR100.
}
\vspace{-0.1cm}
\setlength{\tabcolsep}{5.5mm}{
\renewcommand{\arraystretch}{1.}
\begin{tabular}{|c|c|c|c|c|}
\hline
Method    & \multicolumn{2}{|c|}{Same architecture} & \multicolumn{2}{c|}{Different architecture} \\ \cline{1-5} 
teacher   & vgg13           & resnet32x4           & resnet56            & vgg13                \\
student   & vgg8            & resnet8x4            & vgg8                & resnet8x4            \\ \hline
teacher   & 75.56           & 79.82                & 73.03               & 75.56                \\
student   & 70.29           & 72.34                & 70.29               & 72.34                \\ \hline
KD        & 73.70(0.26)           & 74.39(0.37)           & 73.90(0.32)           & 74.14(0.36)                \\
CRD       & 73.55(0.23)           & 75.11(0.34)           & 73.92(0.40)           & 74.34(0.35)                \\
DynamicKD & \textbf{73.92(0.30)}  & \textbf{76.00(0.27)}  & \textbf{74.51(0.40)}  & \textbf{76.19(0.18)}       \\ \hline
\end{tabular}}
\label{table6}
\end{table}
\vspace{-0.2cm}

The first row of the table indicates whether the teacher and student networks belong to the same type of network architecture. The second and third rows display the teacher and student networks used. The fourth and fifth rows indicate the classification accuracy of the teacher and the student networks obtained by cross-validation experiments. The remaining table shows the classification accuracies of the three distillation algorithms. The cross-validation experiments show that the proposed algorithm DynamicKD still obtains excellent performance. In the distillation experiment resnet32x4 $\rightarrow$ resnet8x4, the proposed algorithm improves 1.61 points over the traditional knowledge distillation algorithm KD and 0.89 points over the state-of-the-art knowledge distillation algorithm CRD. In the distillation experiments vgg13 $\rightarrow$ resnet8x4, the proposed algorithm improves 2.05 points over KD and 1.85 points over CRD.
And the standard deviation of the proposed algorithm is similar to other algorithms.
This shows the effectiveness and efficiency of the proposed algorithm.

\subsection{The Performance Analysis of the Proposed Algorithm}

\subsubsection{The Dynamic Entropy Correction Knowledge Distillation with Different Correction Positions}
\begin{figure}[h]
\centering
\includegraphics[scale=0.55]{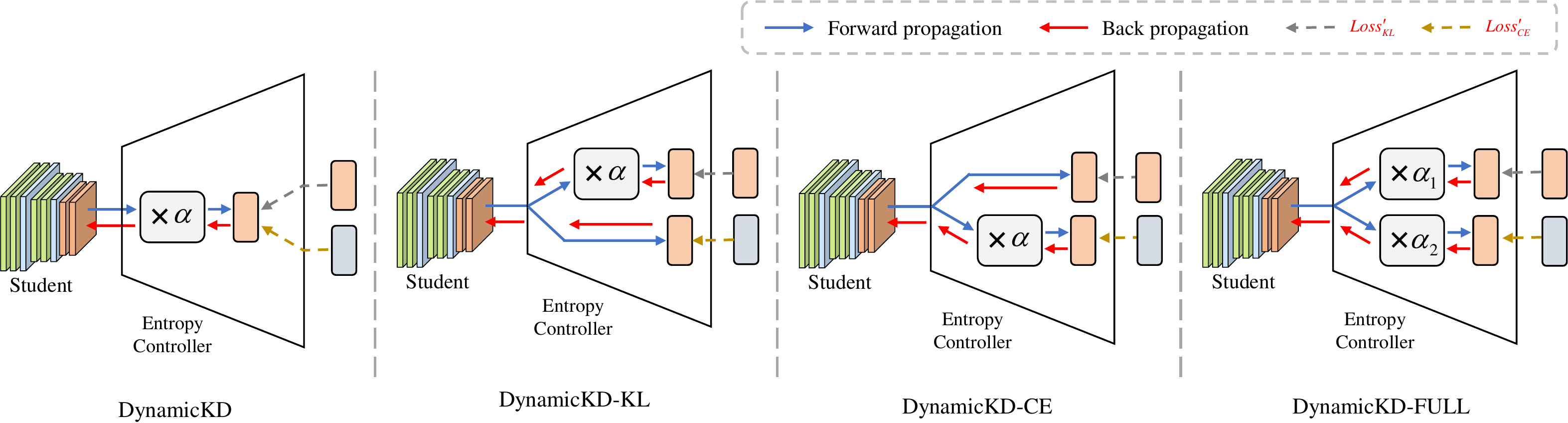}
\caption{DynamicKD and another three dynamic entropy controller structures for the knowledge distillation based on the dynamic entropy correction.}
\label{Fig8}
\end{figure}

This paper proposes a knowledge distillation algorithm that affects KL divergence loss and cross-entropy loss through an entropy controller. This method can improve the training of neural networks. To further understand the effect of the entropy controller on the student networks and the efficiency of the method used in this paper, the corresponding performance analysis experiments are designed. 
This paper designs corresponding performance analysis experiments to investigate further the effect of the dynamic entropy correction on knowledge distillation. 
DynamicKD is intended to improve the distillation performance by correcting the output entropy of the student network during the knowledge distillation. 
Therefore, as shown in Figure \ref{Fig8}, this paper designs another three algorithmic structures by changing the positions of entropy controllers to understand further the entropy controller’s effect on knowledge distillation.
% \vspace{0.2cm}
\vspace{-0.2cm}

\begin{figure}[h]
\centering
\subfloat[DynamicKD-KL\&CE] {\includegraphics[scale=0.10]{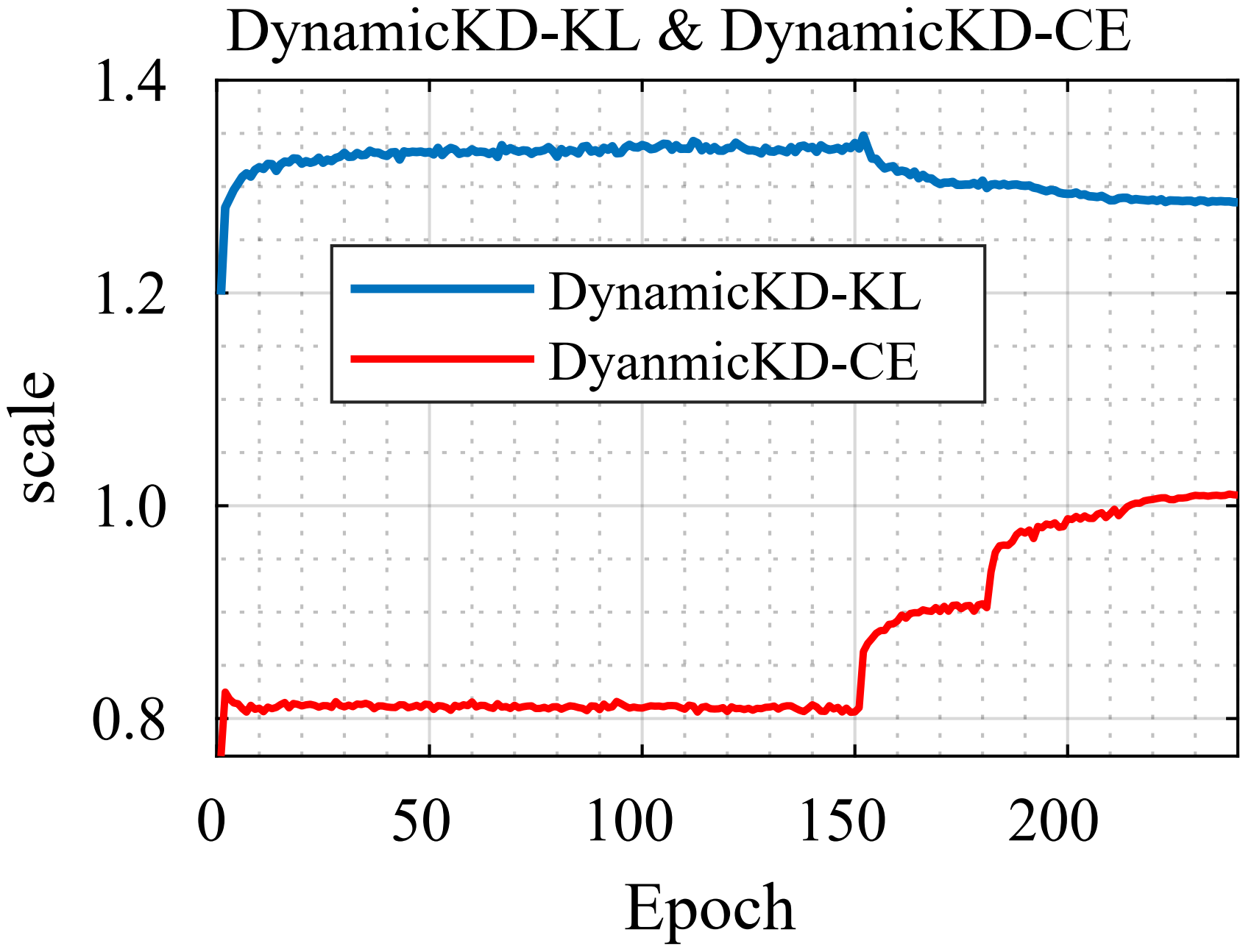}} \hfill
\subfloat[DynamicKD-FULL] {\includegraphics[scale=0.10]{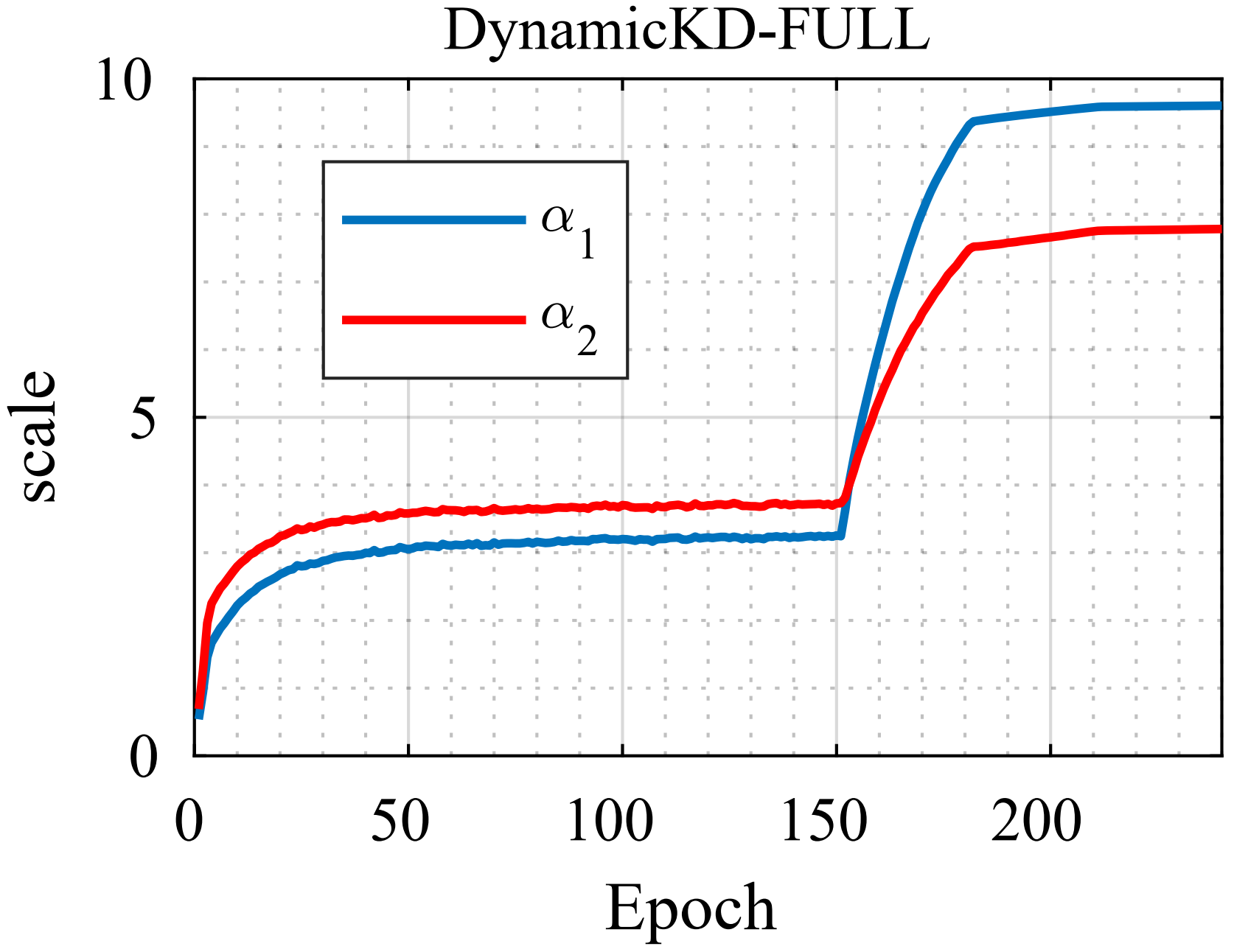}} \hfill
\subfloat[DynamicKD] {\includegraphics[scale=0.10]{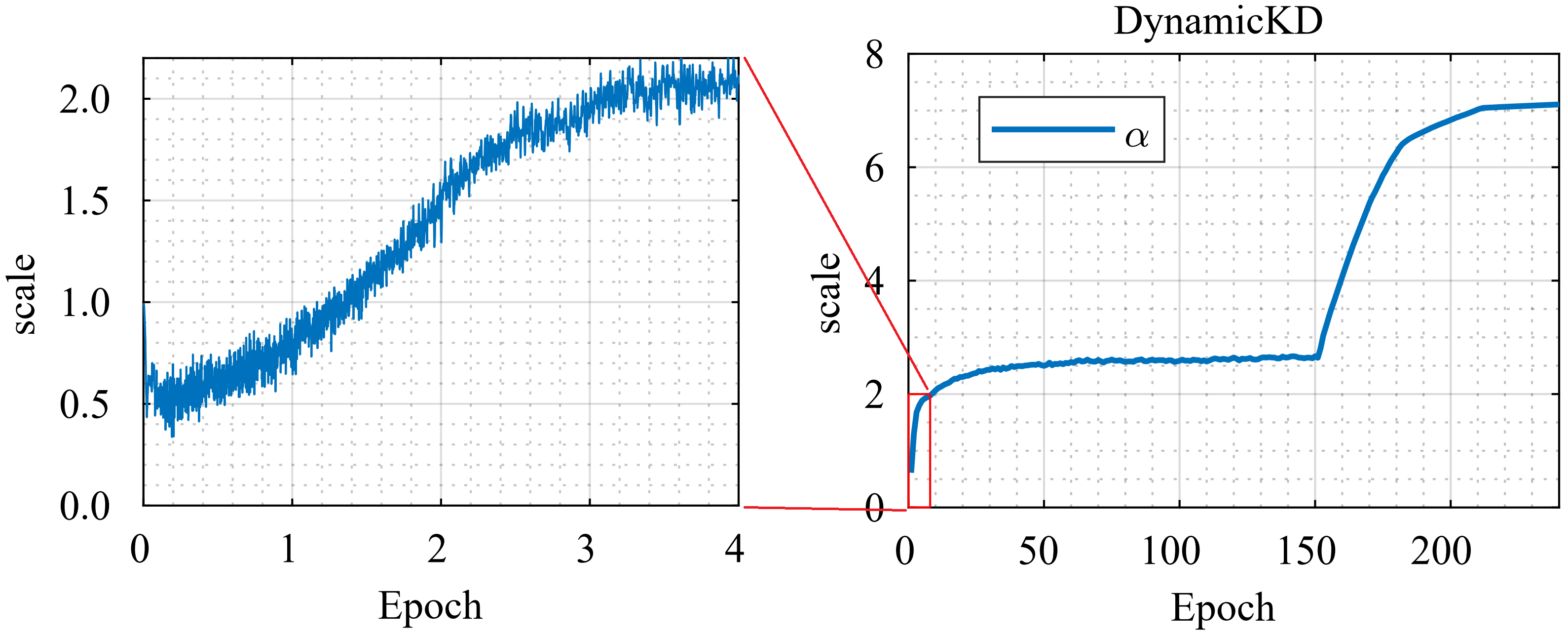}}

\caption{Variation curves of entropy adjustment parameters in DynamicKD-KL, DynamicKD-CE, and DynamicKD-FULL as distillation proceeds.}
\label{Fig9}
\end{figure}

The left entropy controller is the one used in the proposed algorithm, where the same $\alpha$ is for the KL divergence loss and the cross-entropy loss.
The next one only uses the entropy adjustment parameter for the KL divergence loss, denoted as DynamicKD-KL. The following is for the cross-entropy loss only, marked as DynamicKD-CE. The last applies two independent entropy adjustment parameters for both losses, denoted as DynamicKD-FULL. Then, the experiments using DynamicKD, DynamicKD-KL, DynamicKD-CE, and DynamicKD-FULL are conducted on the distillation vgg13 $\rightarrow$ resnet20 with the same experimental setup as Table \ref{table3}.
To further understand the effect of the different dynamic entropy structures on entropy controllers, this paper compares the variation curve of the entropy adjustment parameter in different dynamic entropy controller structures with knowledge distillation. The experiment results are shown in Figure \ref{Fig9}.

\vspace{-0.2cm}
\vspace{-0.2cm}

The results show that
the entropy adjustment parameters of DynamicKD-KL and DynamicKD-CE change less than DynamicKD-FULL and DynamicKD. It may be that these two loss functions interact with each other, and another unadjusted one suppresses the adjusted parameter's change. The change trends of DynamicKD and DynamicKD-FULL are similar. The student network performance is insufficient in the early distillation stage, so the algorithm reduces learning difficulty by applying a lower entropy adjustment parameter. As the knowledge distillation proceeds, the student network performance improves; therefore, the algorithm facilitates the student network training by elevating the adjustment parameter. This phenomenon is consistent with the theoretical analysis in section 2.1. In addition, unlike DynamicKD, the two adjustment parameters of DynamicKD-FULL behave differently at different distillation stages. The parameter for KL divergence loss is more significant in the early stage, and the parameter for cross-entropy loss is more significant in the later stage. A reasonable explanation is that DynamicKD-FULL has flexible adjustability on both losses and can adjust both adjustment parameters independently to reduce knowledge distillation error further.
In addition, this paper compares the performance of these four dynamic entropy-based knowledge distillation algorithms
with the traditional knowledge distillation algorithm KD. The results are shown in Table \ref{table7}, where the teacher is vgg13, and the student is resnet20.
\begin{table}[ht]
\footnotesize
\centering
\caption{
Classification accuracies (\%) of different dynamic entropy-based distillation algorithms on the CIFAR100.
}
\vspace{-0.1cm}
\setlength{\tabcolsep}{2.4mm}{
\renewcommand{\arraystretch}{1}
\begin{tabular}{|c|c|c|c|c|c|c|c|}
\hline
Algorithm & Teacher & Student & KD   & DynamicKD-KL & DynamicKD-CE & DynamicKD-FULL & DynamicKD \\ \hline
Accuracy  & 75.13   & 69.54   & 69.4 & 70.42        & 70.41        & 70.73          & 70.81    \\
\hline
\end{tabular}}
\label{table7}
\end{table}
\vspace{-0.2cm}

As shown in Table \ref{table7},
all four distillation algorithms based on dynamic correction perform better than the traditional knowledge distillation algorithm KD. This phenomenon indicates that reducing the learning difficulty by dynamically correcting the output entropy of the student can enhance the distillation performance. Meanwhile, DynamicKD-KL and DynamicKD-CE slightly underperform among these dynamic correction algorithms, indicating that the single adjustment of KL divergence loss or cross-entropy loss does not optimize the distillation gap well. And simultaneously adjusting both losses is more helpful in improving the distillation performance. In addition, the performance of DynamicKD-FULL is slightly weaker than DynamicKD. A reasonable explanation is that DynamicKD-FULL can reduce distillation losses better than DynamicKD, but this excessive reduction does not improve overall performance. It is like overfitting, while the synchronous adjustment of both losses can reduce this overfitting-like effect and enhance the performance of knowledge distillation.

\subsubsection{The Knowledge distillation with segmented static entropy correction}
The above experiments show that dynamically adjusting the output entropy of the student network can optimize knowledge distillation. But, can the same effect is achieved by designing a simple segmental adjustment strategy similar to the changing trend of parameter $\alpha$? Inspired by this problem, this paper designed a simple segmented entropy correction method. It divides the whole training process into three training periods, the early period (0-10 Epoch), the middle period (10-120 Epoch), and the last period (120-240 Epoch). It adjusts the output entropy during each period using the parameter $\alpha$ with the method defined by Equation (\ref{eq4}). 
Inspired by the experiment in subsection 3.5.1, in the early period, the student network is insufficient, so a lower parameter $\alpha$ (0.5) is used to reduce the learning difficulty. As the distillation proceeds, the student network performance improves in the middle period, so a parameter $\alpha$ (1.0) is chosen. In the last period, the performance improves further, so a higher parameter $\alpha$ (2.0) is applied for better learning.
This paper conducts experiments on CIFAR100. The teachers are resnet32x4 and vgg13, the student is resnet8x4, and the experimental results are shown in Table \ref{table8}.
% \vspace{0.2cm}
\begin{table}[h]
\footnotesize
\centering
\caption{Classification accuracies (\%) of KD and KD-static on CIFAR100.}
\vspace{-0.1cm}
\setlength{\tabcolsep}{6mm}{
\renewcommand{\arraystretch}{1.1}
\begin{tabular}{|c|c|c|}
\hline
Method    & resnet32x4 $\rightarrow$ resnet8x4     & vgg13 $\rightarrow$ resnet8x4 \\ \hline
teacher   & 79.24          & 75.13            \\
student   & 72.85          & 72.85            \\ \hline
KD        & 73.42          & 73.74            \\
KD-static & 74.59          & 74.99            \\
DynamicKD & \textbf{76.06} & \textbf{76.01}   \\ \hline
\end{tabular}}
\label{table8}
\end{table}
\vspace{-0.2cm}

As shown in Table \ref{table8}, KD denotes the traditional knowledge distillation algorithm, and KD-static represents the knowledge distillation using the simple segmented entropy correction method mentioned above.
In the distillation experiment resnet32x4 $\rightarrow$ resnet8x4, KD-static improves 1.17 points over KD. In the distillation experiment vgg13 $\rightarrow$ resnet8x4, KD-static improves 1.25 points over KD. The simple adjusting method obtains more than 1 point accuracy improvement, demonstrating the value of this simple strategy for knowledge distillation.
But compared to DynamicKD, the improvement brought by this simple strategy is still limited. In the distillation experiments resnet32x4 $\rightarrow$ resnet8x4 and vgg13 $\rightarrow$ resnet8x4, DynamicKD improves 1.47 and 1.02 points over this simple strategy, respectively.
This demonstrates the effectiveness and efficiency of the proposed algorithm.

\subsubsection{The Dynamic Entropy Correction Knowledge Distillation with Different Performance Teachers}
This paper proposes a knowledge distillation algorithm to reduce student learning difficulty and improve performance.
However, the effect of teacher networks with different performances on the distillation performance is not negligible. Therefore, this section investigates the effect of teacher networks with different performances on the proposed algorithm. 

\noindent\textit{(1) The Ablation Experiments with Teacher Networks of Different Sizes}

\begin{figure}[h]
\centering
\includegraphics[scale=0.6]{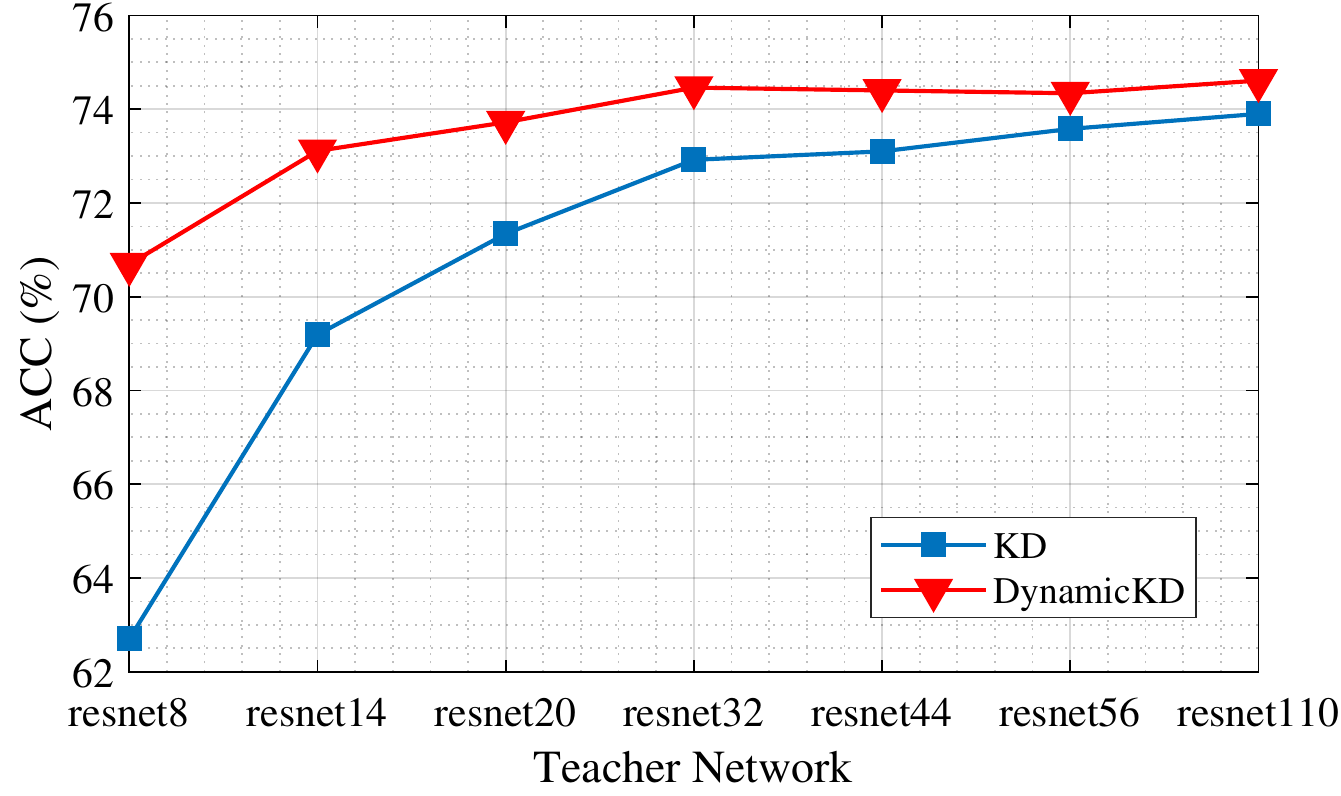}
\caption{The distillation experiments using teacher networks with different sizes.}
\label{Fig11}
\end{figure}

The size of the neural network affects the neural network's performance. So this paper first conducts the distillation experiments using teachers of different sizes.
The teachers used are resnet8, resnet14, resnet20, resnet32, resnet44, resnet56 and resnet110. 
Figure \ref{Fig11} shows the experiment results.

\vspace{-0.2cm}

The teacher networks are ResNet-like networks with different sizes, and the student network is vgg8. The larger the teacher network is, the stronger its performance is. From the figure, both the traditional knowledge distillation KD and the proposed algorithm DynamicKD show a similar growth trend in distillation performance as the size of the teacher network increases. The difference is that DynamicKD performs better in all distillation experiments and has fewer performance fluctuations. This shows the advantage of the proposed algorithm, which can dynamically adjust the output of the student network so that the student network performs well in different distillation experiments.

\noindent\textit{(2) The Distillation Experiments on Teacher Networks with Different Performance Weakening}

Research works have found that reducing the teacher’s performance can improve distillation performance \cite{cho_efficacy_2019}. 
Therefore, this paper conducts experiments on knowledge distillation to investigate the relationship between distillation algorithms by weakening the teacher network and the proposed algorithm.
The experiments are conducted on the CIFAR100 with DynamicKD and the traditional knowledge distillation algorithm KD, and the experimental results are shown in Table \ref{table9}. The teacher network is trained in the same way as the traditional teacher network except for the training epochs. 
\begin{table}[h]
\footnotesize
\centering
\caption{
classification accuracies (\%) with different weakening teachers on CIFAR100.
}
\vspace{-0.1cm}
\setlength{\tabcolsep}{3.5mm}{
\renewcommand{\arraystretch}{1}
\begin{tabular}{|c|c|c|c|c|c|}
\hline
\multicolumn{2}{|c|}{\multirow{2}{*}{Method}}        & \multicolumn{4}{c|}{Teacher-epochs} \\ \cline{3-6} 
\multicolumn{2}{|c|}{}                                           & 80     & 170    & 200   & 240   \\ \hline
\multirow{2}{*}{resnet32x4 $\rightarrow$ resnet8x4} & DynamicKD & 75.37  & \textbf{76.75}  & 76.21 & 76.06 \\
                                                    & KD        & 67.66  & \textbf{74.07}  & 74.01 & 73.42 \\ \hline
\multirow{2}{*}{vgg13 $\rightarrow$ resnet8x4}      & DynamicKD & 74.09  & \textbf{76.18}  & 75.90 & 76.01 \\
                                                    & KD        & 64.57  & 73.51  & 73.73 & \textbf{73.74} \\ \hline
\multirow{2}{*}{vgg13 $\rightarrow$ vgg8}           & DynamicKD & 72.65  & 74.12  & \textbf{74.19} & 74.13 \\
                                                    & KD        & 66.35  & 73.35  & \textbf{73.80} & 73.44 \\ \hline
\end{tabular}}
\label{table9}
\end{table}
\vspace{-0.2cm}

The experiments show that appropriately weakening the teacher could improve the distillation performance, the same as the published experimental findings \cite{cho_efficacy_2019}. In the distillation experiment resnet32x4 $\rightarrow$ resnet8x4, DynamicKD and KD obtained 76.06\% and 73.42\% accuracy with better teachers (epoch=240), while they obtained better accuracy (76.75\% and 74.07\%) with lower teachers (epoch=170).
But inappropriately weakening the teacher can also lead to decreased distillation performance. In the distillation experiment vgg13 $\rightarrow$ resnet8x4, the low-performance teacher (epoch=80) reduced the performance of DynamicKD and KD from the high-performance teacher (epoch=240) by 1.92 and 9.17 points, respectively. This indicates that the method of weakening teachers needs to be set reasonably for the weakening parameter. And it is also found from these three distillation experiments that the appropriate setting varies with the distillation experiment. The setting of this parameter undoubtedly increases the difficulty of using the weakening method.
In addition, these two methods reduce the learning difficulty of the student network in different ways. The weakening method is from the teacher, while the proposed algorithm is from the student. 
The experiments also found that these two methods are not conflicting but complementary. Appropriately weakening the performance of the teacher network can also improve the performance of the proposed algorithm. In all three distillation experiments, weakening the teacher (epoch=170 or 200) can improve the performance of the original DynamicKD (epoch=240).

\noindent\textit{(3) The Distillation Experiments on Teacher Networks with Entropy Controller}

This paper conducts experiments to change the output entropy of the teacher with the entropy controller. It adds the entropy controller for the teacher instead of the student, named DynamicKD-Teacher. The experimental results are shown below, where the teacher is vgg13 and the student is resnet8x4.
\begin{table}[h]
\footnotesize
\centering
\caption{
Classification accuracies (\%) of KD, DynamicKD, DynamicKD-Teacher on CIFAR100.
}
\vspace{-0.1cm}
\setlength{\tabcolsep}{3.5mm}{
\renewcommand{\arraystretch}{1}
\begin{tabular}{|c|c|c|c|c|c|}
\hline
teacher & student & KD    & DynamicKD   & DynamicKD-Teacher  \\ \hline
75.13   & 72.85   & 73.74 & 76.01       & 73.53              \\ \hline 

\end{tabular}}
\label{table10}
\end{table}
\vspace{-0.2cm}

The experimental results show that adding only the entropy controller to the teacher brings a performance drop of 0.21 points compared to the performance of KD without the entropy controller. It was also found that the $\alpha$ on the entropy controller from the teacher remained relatively low during the whole distillation process, below 0.5. This suggests that the entropy controller has been reducing the learning difficulty of the knowledge generated by the teacher. One possible explanation is that it is more beneficial for teachers to produce knowledge easier to learn to reduce the gap between teacher and student. However, in the later stages of training, this knowledge may hinder the improvement of student performance.

\subsubsection{
The experiments about learnable distillation temperature and balancing $Loss_{KL}$ and $Loss_{CE}$ 
}
Compared to KD, DynamicKD adjusts the output entropy of the student network through an entropy controller. However, the distillation temperature $T$ in KD also adjusts the output entropy. Adjusting the output entropy of the student by a learnable distillation temperature $T$ may obtain the same effect. In KD, the distillation temperature affects the KL divergence loss, which needs to multiply the loss by $T^2$ to make the KL divergence loss have the same scale as the cross-entropy loss. For DynamicKD, the entropy controller affects both the KL divergence loss and the cross-entropy loss, and both losses still have the same scale. However, how does multiplying the loss by $\frac{1}{\alpha ^2}$ affect the algorithm’s performance? To this end, the DynamicKD-compensation and DynamicKD-T algorithms are designed. DynamicKD-compensation’s loss function is scaled by $\frac{1}{\alpha ^2}$ compared to DynamicKD. DynamicKD-T trains the student with a learnable temperature $T$ compared to KD. The experiment result are shown in Table \ref{table11}, where the teacher is vgg13 and the student is resnet8x4.

% \vspace{0.2cm}
\begin{table}[h]
\footnotesize
\centering
\caption{
Classification accuracies (\%) of KD, DynamicKD, DynamicKD-T and DynamicKD-compensation on CIFAR100.
}
\vspace{-0.1cm}
\setlength{\tabcolsep}{3.5mm}{
\renewcommand{\arraystretch}{1.1}
\begin{tabular}{|c|c|c|c|c|c|c|}
\hline
teacher & student & KD    & DynamicKD   & DynamicKD-compensation  & DynamicKD-T \\ \hline
75.13   & 72.85   & 73.74 & 76.01       & 73.16                   & 74.80       \\ \hline 
\end{tabular}}
\label{table11}
\end{table}
\vspace{-0.2cm}

The experimental results show that DynamicKD-compensation does not obtain a better performance than DynamicKD; it has a lower performance than KD. For both KD and DynamicKD, it is important that the KL divergence loss and the cross entropy loss have the same scale. Breaking the balance can affect distillation performance.
DynamicKD-T performs better than KD. This suggests that setting a learnable distillation temperature $T$ for the student network can also reduce the training difficulty and improve performance. However, it did not perform as well as DynamicKD. Distillation temperature only affects KL divergence loss. Similar to the experiment in section 3.6.1, the adjustment only for KL divergence loss would limit its optimization ability.

\subsubsection{The Computational Time Analysis of the Proposed Algorithm}
To further analyze the efficiency of the proposed algorithm, this paper also compares the running time of the proposed algorithm and the traditional distillation algorithm KD in several distillation experiments. The experimental results are shown in Table \ref{table12}.

% \vspace{0.2cm}
\begin{table}[h]
\footnotesize
\centering
\caption{The running time of the DynamicKD and KD on CIFAR100 and ImageNet. 
}
\vspace{-0.1cm}
\setlength{\tabcolsep}{5mm}{
\renewcommand{\arraystretch}{1}
\begin{tabular}{|c|c|c|c|c|c|}
\hline
DataSet                   & Teacher                   & Student   & KD                & DynamicKD         & cost $\uparrow$ \\ \hline
\multirow{7}{*}{CIFAR100} & resnet110                 & resnet20  & 1 hour 25 minutes & 1 hour 22 minutes & -4\%            \\ \cline{2-6} 
                          & resnet32x4                & resnet8x4 & 53 minutes        & 54 minutes        & 2\%             \\ \cline{2-6} 
                          & \multirow{2}{*}{resnet56} & resnet20  & 1 hour 1 minute   & 1 hour 3 minutes  & 3\%             \\
                          &                           & vgg8      & 46 minutes        & 46 minutes        & 0\%             \\ \cline{2-6} 
                          & \multirow{3}{*}{vgg13}    & resnet20  & 41 minutes        & 42 minutes        & 2\%             \\
                          &                           & resnet8x4 & 34 minutes        & 35 minutes        & 3\%             \\
                          &                           & vgg8      & 26 minutes        & 27 minutes        & 4\%             \\ \hline
ImageNet                  & ResNet34                  & ResNet18  & 2 day 15 hours    & 2 day 14 hours    & -1\%            \\ \hline
\end{tabular}}
\label{table12}
\end{table}
\vspace{-0.2cm}

The last column of the table shows the consumed time increase in DynamicKD compared to KD.
The experiments show no significant difference between the running time of DynamicKD and KD. This demonstrates the efficiency of the proposed algorithm. It does not significantly increase computational complexity while improving the distillation performance.
It only increases the running time by 0-4\%. It even shows a 1\% and 4\% time reduction in the ResNet34$\rightarrow$ResNet18 and resnet110$\rightarrow$resnet20 distillation experiments, which may be due to the optimization of the deep learning library. These results indicate that the resource consumption of the proposed algorithm is negligible.

\section{Conclusion and Future Work}
During the knowledge distillation, it is challenging for the lightweight student network to directly learn the knowledge from the high-performance teacher network. The distillation gap can hinder student training. To this end, this paper proposes a knowledge distillation algorithm based on dynamic entropy correction, which uses an entropy controller to adjust the output entropy of the student network adaptively. This approach can reduce the distillation gaps and improve the distillation performance. The proposed algorithm’s performance is evaluated in various knowledge distillation experiments, and the comparisons with many state-of-the-art knowledge distillation algorithms demonstrate the effectiveness and efficiency of the proposed algorithm. In particular, on the CIFAR100 benchmark dataset, the proposed algorithm improves the advanced algorithm CRD from 75.19\% to 76.06\% in teacher-student pair resnet32x4-resnet8x4, and it improves 2.64 points than the traditional knowledge distillation KD.

The distillation gaps always exit and change during the distillation. This work demonstrates that reducing the gap from the student network is possible. However, this paper only employs a simple entropy adjustment strategy, and there is still much room for research. There may be other better ways to reduce the distillation gaps. In future work, other more advanced methods should be considered. In addition, the interaction of the entropy controller with other distillation algorithms will be investigated.

\section*{Acknowledgements}
This work was partially supported by the National Natural Science Foundation of China under Grants Nos. 62176200 and 61871306, 
the National Key R\&D Program of China and the Guangdong Provincial Key Laboratory under Grant No. 2020B121201001, 
the Natural Science Basic Research Program of Shaanxi under Grant No.2022JC-45, 2022JQ-616,
the Open Research Projects of Zhejiang Lab under Grant 2021KG0AB03.

\bibliographystyle{elsarticle-num} 
\setlength{\bibsep}{0em}
% \begin{spacing}{0.0}
\bibliography{mybib}

\begin{thebibliography}{10}
\expandafter\ifx\csname url\endcsname\relax
  \def\url#1{\texttt{#1}}\fi
\expandafter\ifx\csname urlprefix\endcsname\relax\def\urlprefix{URL }\fi
\expandafter\ifx\csname href\endcsname\relax
  \def\href#1#2{#2} \def\path#1{#1}\fi

\bibitem{peng_few-shot_2019}
Z.~Peng, Z.~Li, J.~Zhang, Y.~Li, G.-J. Qi, J.~Tang, Few-{Shot} {Image}
  {Recognition} {With} {Knowledge} {Transfer}, Proceedings of the {IEEE}/{CVF}
  {International} {Conference} on {Computer} {Vision} (2019) 441--449.

\bibitem{li_ctnet_2022}
Z.~Li, Y.~Sun, L.~Zhang, J.~Tang, {CTNet}: {Context}-{Based} {Tandem} {Network}
  for {Semantic} {Segmentation}, IEEE Transactions on Pattern Analysis and
  Machine Intelligence 44~(12) (2022) 9904--9917.

\bibitem{yu_learning_2015}
J.~Yu, D.~Tao, M.~Wang, Y.~Rui, Learning to {Rank} {Using} {User} {Clicks} and
  {Visual} {Features} for {Image} {Retrieval}, IEEE Transactions on Cybernetics
  45~(4) (2015) 767--779.

\bibitem{li_deep_2019}
Z.~Li, J.~Tang, T.~Mei, Deep {Collaborative} {Embedding} for {Social} {Image}
  {Understanding}, IEEE Transactions on Pattern Analysis and Machine
  Intelligence 41~(9) (2019) 2070--2083.

\bibitem{yu_hierarchical_2022}
J.~Yu, M.~Tan, H.~Zhang, Y.~Rui, D.~Tao, Hierarchical {Deep} {Click} {Feature}
  {Prediction} for {Fine}-{Grained} {Image} {Recognition}, IEEE Transactions on
  Pattern Analysis and Machine Intelligence 44~(2) (2022) 563--578.

\bibitem{hong_multimodal_2019}
C.~Hong, J.~Yu, J.~Zhang, X.~Jin, K.-H. Lee, Multimodal {Face}-{Pose}
  {Estimation} {With} {Multitask} {Manifold} {Deep} {Learning}, IEEE
  Transactions on Industrial Informatics 15~(7) (2019) 3952--3961.

\bibitem{hong_multimodal_2015}
C.~Hong, J.~Yu, J.~Wan, D.~Tao, M.~Wang, Multimodal {Deep} {Autoencoder} for
  {Human} {Pose} {Recovery}, IEEE Transactions on Image Processing 24~(12)
  (2015) 5659--5670.

\bibitem{hong_image-based_2015}
C.~Hong, J.~Yu, D.~Tao, M.~Wang, Image-{Based} {Three}-{Dimensional} {Human}
  {Pose} {Recovery} by {Multiview} {Locality}-{Sensitive} {Sparse} {Retrieval},
  IEEE Transactions on Industrial Electronics 62~(6) (2015) 3742--3751.

\bibitem{szegedy_going_2015}
C.~Szegedy, W.~Liu, Y.~Jia, P.~Sermanet, S.~Reed, D.~Anguelov, D.~Erhan,
  V.~Vanhoucke, A.~Rabinovich, Going deeper with convolutions, Proceedings of
  the {IEEE} conference on computer vision and pattern recognition (2015) 1--9.

\bibitem{he_deep_2016}
K.~He, X.~Zhang, S.~Ren, J.~Sun, Deep {Residual} {Learning} for {Image}
  {Recognition}, Proceedings of the {IEEE} {Conference} on {Computer} {Vision}
  and {Pattern} {Recognition} (2016) 770--778.

\bibitem{zhang_semi-supervised_2021}
M.~Zhang, Y.~Zhou, J.~Zhao, S.~Xia, J.~Wang, Z.~Huang, Semi-supervised
  blockwisely architecture search for efficient lightweight generative
  adversarial network, Pattern Recognition 112 (2021) 107794.

\bibitem{shang_evolutionary_2022}
R.~Shang, S.~Zhu, J.~Ren, H.~Liu, L.~Jiao, Evolutionary neural architecture
  search based on evaluation correction and functional units, Knowledge-Based
  Systems 251 (2022) 109206.

\bibitem{sandler_mobilenetv2_2018}
M.~Sandler, A.~Howard, M.~Zhu, A.~Zhmoginov, L.~C. Chen, {MobileNetV2}:
  {Inverted} {Residuals} and {Linear} {Bottlenecks}, Proceedings of the {IEEE}
  {Conference} on {Computer} {Vision} and {Pattern} {Recognition} (2018)
  4510--4520.

\bibitem{cheng_survey_2020}
Y.~Cheng, D.~Wang, P.~Zhou, T.~Zhang, A {Survey} of {Model} {Compression} and
  {Acceleration} for {Deep} {Neural} {Networks}, arXiv:1710.09282 (2020).

\bibitem{yao_deep_2021}
K.~Yao, F.~Cao, Y.~Leung, J.~Liang, Deep neural network compression through
  interpretability-based filter pruning, Pattern Recognition 119 (2021) 108056.

\bibitem{mirzadeh_improved_2020}
S.~I. Mirzadeh, M.~Farajtabar, A.~Li, N.~Levine, A.~Matsukawa, H.~Ghasemzadeh,
  Improved {Knowledge} {Distillation} via {Teacher} {Assistant}, Proceedings of
  the AAAI Conference on Artificial Intelligence 34~(04) (2020) 5191--5198.

\bibitem{chen_shallowing_2019}
S.~Chen, Q.~Zhao, Shallowing {Deep} {Networks}: {Layer}-{Wise} {Pruning}
  {Based} on {Feature} {Representations}, IEEE Transactions on Pattern Analysis
  and Machine Intelligence 41~(12) (2019) 3048--3056.

\bibitem{guo_dmcp_2020}
S.~Guo, Y.~Wang, Q.~Li, J.~Yan, {DMCP}: {Differentiable} {Markov} {Channel}
  {Pruning} for {Neural} {Networks}, Proceedings of the {IEEE}/{CVF}
  {Conference} on {Computer} {Vision} and {Pattern} {Recognition} (2020)
  1539--1547.

\bibitem{he_asymptotic_2020}
Y.~He, X.~Dong, G.~Kang, Y.~Fu, C.~Yan, Y.~Yang, Asymptotic {Soft} {Filter}
  {Pruning} for {Deep} {Convolutional} {Neural} {Networks}, IEEE Transactions
  on Cybernetics 50~(8) (2020) 3594--3604.

\bibitem{hinton_distilling_2015}
G.~Hinton, O.~Vinyals, J.~Dean, Distilling the knowledge in a neural network,
  arXiv:1503.02531 (2015).

\bibitem{bucilua_model_2006}
C.~Buciluǎ, R.~Caruana, A.~Niculescu-Mizil, Model compression, Proceedings of
  the 12th {ACM} {SIGKDD} international conference on {Knowledge} discovery and
  data mining (2006) 535--541.

\bibitem{10.1145/3487042}
Y.~Pan, Z.~Li, L.~Zhang, J.~Tang, Causal inference with knowledge distilling
  and curriculum learning for unbiased vqa, ACM Trans. Multimedia Comput.
  Commun. Appl. 18~(3) (2022).

\bibitem{wang_joint_2021}
Z.~R. Wang, J.~Du, Joint architecture and knowledge distillation in {CNN} for
  {Chinese} text recognition, Pattern Recognition 111 (2021) 107722.

\bibitem{shang_hyperspectral_2022}
R.~Shang, J.~Ren, S.~Zhu, W.~Zhang, J.~Feng, Y.~Li, L.~Jiao, Hyperspectral
  {Image} {Classification} {Based} on {Pyramid} {Coordinate} {Attention} and
  {Weighted} {Self}-{Distillation}, IEEE Transactions on Geoscience and Remote
  Sensing 60 (2022) 1--16.

\bibitem{li_hierarchical_2021}
W.~Li, S.~Gong, X.~Zhu, Hierarchical distillation learning for scalable person
  search, Pattern Recognition 114 (2021) 107862.

\bibitem{romero_fitnets_2015}
A.~Romero, N.~Ballas, S.~E. Kahou, A.~Chassang, C.~Gatta, Y.~Bengio, Fitnets:
  {Hints} for thin deep nets, International {Conference} on {Learning}
  {Representations} (2015).

\bibitem{komodakis_paying_2017}
S.~Zagoruyko, N.~Komodakis, Paying more attention to attention: improving the
  performance of convolutional neural networks via attention transfer,
  International {Conference} on {Learning} {Representations} (2017).

\bibitem{ahn_variational_2019}
S.~Ahn, S.~X. Hu, A.~Damianou, N.~D. Lawrence, Z.~Dai, Variational
  {Information} {Distillation} for {Knowledge} {Transfer}, 2019 {IEEE}/{CVF}
  {Conference} on {Computer} {Vision} and {Pattern} {Recognition} ({CVPR})
  (2019) 9155--9163.

\bibitem{yim_gift_2017}
J.~Yim, D.~Joo, J.~Bae, J.~Kim, A {Gift} from {Knowledge} {Distillation}:
  {Fast} {Optimization}, {Network} {Minimization} and {Transfer} {Learning},
  2017 {IEEE} {Conference} on {Computer} {Vision} and {Pattern} {Recognition}
  ({CVPR}) (2017) 4133--4141.

\bibitem{cho_efficacy_2019}
J.~H. Cho, B.~Hariharan, On the {Efficacy} of {Knowledge} {Distillation}, 2019
  {IEEE}/{CVF} {International} {Conference} on {Computer} {Vision} ({ICCV})
  (2019) 4793--4801.

\bibitem{zhao_highlight_2020}
H.~Zhao, X.~Sun, J.~Dong, C.~Chen, Z.~Dong, Highlight {Every} {Step}:
  {Knowledge} {Distillation} via {Collaborative} {Teaching}, IEEE Transactions
  on Cybernetics (2020) 1--12.

\bibitem{jin_knowledge_2019}
X.~Jin, B.~Peng, Y.~Wu, Y.~Liu, J.~Liu, D.~Liang, J.~Yan, X.~Hu, Knowledge
  {Distillation} via {Route} {Constrained} {Optimization}, Proceedings of the
  {IEEE}/{CVF} {International} {Conference} on {Computer} {Vision} (2019)
  1345--1354.

\bibitem{zhang_deep_2018}
Y.~Zhang, T.~Xiang, T.~M. Hospedales, H.~Lu, Deep {Mutual} {Learning},
  Proceedings of the {IEEE} {Conference} on {Computer} {Vision} and {Pattern}
  {Recognition} (2018) 4320--4328.

\bibitem{lan_knowledge_2018}
X.~Lan, X.~Zhu, S.~Gong, Knowledge distillation by on-the-fly native ensemble,
  Proceedings of the 32nd {International} {Conference} on {Neural}
  {Information} {Processing} {Systems} (2018) 7528--7538.

\bibitem{guo_online_2020}
Q.~Guo, X.~Wang, Y.~Wu, Z.~Yu, D.~Liang, X.~Hu, P.~Luo, Online {Knowledge}
  {Distillation} via {Collaborative} {Learning}, 2020 {IEEE}/{CVF} {Conference}
  on {Computer} {Vision} and {Pattern} {Recognition} ({CVPR}) (2020)
  11017--11026.

\bibitem{wu_peer_2021}
G.~Wu, S.~Gong, Peer {Collaborative} {Learning} for {Online} {Knowledge}
  {Distillation}, Proceedings of the AAAI Conference on Artificial Intelligence
  35~(12) (2021) 10302--10310.

\bibitem{yun_regularizing_2020}
S.~Yun, J.~Park, K.~Lee, J.~Shin, Regularizing {Class}-{Wise} {Predictions} via
  {Self}-{Knowledge} {Distillation}, 2020 {IEEE}/{CVF} {Conference} on
  {Computer} {Vision} and {Pattern} {Recognition} ({CVPR}) (2020) 13873--13882.

\bibitem{grandvalet_semi-supervised_2004}
Y.~Grandvalet, Y.~Bengio, Semi-supervised learning by entropy minimization,
  Advances in Neural Information Processing Systems 17 (2004).

\bibitem{vu_advent_2019}
T.-H. Vu, H.~Jain, M.~Bucher, M.~Cord, P.~Perez, {ADVENT}: {Adversarial}
  {Entropy} {Minimization} for {Domain} {Adaptation} in {Semantic}
  {Segmentation}, Proceedings of the {IEEE}/{CVF} {Conference} on {Computer}
  {Vision} and {Pattern} {Recognition} (2019) 2517--2526.

\bibitem{chen_domain_2019}
M.~Chen, H.~Xue, D.~Cai, Domain {Adaptation} for {Semantic} {Segmentation}
  {With} {Maximum} {Squares} {Loss}, Proceedings of the {IEEE}/{CVF}
  {International} {Conference} on {Computer} {Vision} (2019) 2090--2099.

\bibitem{xu_larger_2019}
R.~Xu, G.~Li, J.~Yang, L.~Lin, Larger {Norm} {More} {Transferable}: {An}
  {Adaptive} {Feature} {Norm} {Approach} for {Unsupervised} {Domain}
  {Adaptation}, Proceedings of the {IEEE}/{CVF} {International} {Conference} on
  {Computer} {Vision} (2019) 1426--1435.

\bibitem{krizhevsky_learning_2009}
A.~Krizhevsky, G.~Hinton, Learning multiple layers of features from tiny
  images, Master's thesis, Department of Computer Science, University of
  Toronto (2009).

\bibitem{deng_imagenet_2009}
J.~Deng, W.~Dong, R.~Socher, L.~J. Li, K.~Li, L.~Fei~Fei, {ImageNet}: {A}
  large-scale hierarchical image database, 2009 {IEEE} {Conference} on
  {Computer} {Vision} and {Pattern} {Recognition} (2009) 248--255.

\bibitem{park_relational_2019}
W.~Park, D.~Kim, Y.~Lu, M.~Cho, Relational {Knowledge} {Distillation}, 2019
  {IEEE}/{CVF} {Conference} on {Computer} {Vision} and {Pattern} {Recognition}
  ({CVPR}) (2019) 3962--3971.

\bibitem{passalis_learning_2018}
N.~Passalis, A.~Tefas, Learning {Deep} {Representations} with {Probabilistic}
  {Knowledge} {Transfer}, Proceedings of the {European} {Conference} on
  {Computer} {Vision} ({ECCV}) (2018) 268--284.

\bibitem{tian_contrastive_2020}
Y.~Tian, D.~Krishnan, P.~Isola, Contrastive {Representation} {Distillation},
  International {Conference} on {Learning} {Representations} (2020).

\bibitem{zhou_rethinking_2021}
H.~Zhou, L.~Song, J.~Chen, Y.~Zhou, G.~Wang, J.~Yuan, Q.~Zhang, Rethinking
  {Soft} {Labels} for {Knowledge} {Distillation}: {A} {Bias}–{Variance}
  {Tradeoff} {Perspective}, International {Conference} on {Learning}
  {Representations} (2021).

\bibitem{huang_like_2019}
Z.~Huang, N.~Wang, Like {What} {You} {Like}: {Knowledge} {Distill} via {Neuron}
  {Selectivity} {Transfer}, International {Conference} on {Learning}
  {Representations} (2019).

\bibitem{heo_comprehensive_2019}
B.~Heo, J.~Kim, S.~Yun, H.~Park, N.~Kwak, J.~Y. Choi, A {Comprehensive}
  {Overhaul} of {Feature} {Distillation}, Proceedings of the {IEEE}/{CVF}
  {International} {Conference} on {Computer} {Vision} (2019) 1921--1930.

\bibitem{tung_similarity-preserving_2019}
F.~Tung, G.~Mori, Similarity-{Preserving} {Knowledge} {Distillation},
  Proceedings of the {IEEE}/{CVF} {International} {Conference} on {Computer}
  {Vision} (2019) 1365--1374.

\bibitem{peng_correlation_2019}
B.~Peng, X.~Jin, J.~Liu, D.~Li, Y.~Wu, Y.~Liu, S.~Zhou, Z.~Zhang, Correlation
  {Congruence} for {Knowledge} {Distillation}, Proceedings of the {IEEE}/{CVF}
  {International} {Conference} on {Computer} {Vision} (2019) 5007--5016.

\bibitem{heo_knowledge_2019}
B.~Heo, M.~Lee, S.~Yun, J.~Y. Choi, Knowledge {Transfer} via {Distillation} of
  {Activation} {Boundaries} {Formed} by {Hidden} {Neurons}, Proceedings of the
  AAAI Conference on Artificial Intelligence 33~(01) (2019) 3779--3787.

\bibitem{kim_paraphrasing_2018}
J.~Kim, S.~Park, N.~Kwak, Paraphrasing complex network: network compression via
  factor transfer, Proceedings of the 32nd {International} {Conference} on
  {Neural} {Information} {Processing} {Systems} (2018) 2765--2774.

\bibitem{chen_distilling_2021}
P.~Chen, S.~Liu, H.~Zhao, J.~Jia, Distilling {Knowledge} via {Knowledge}
  {Review}, Proceedings of the {IEEE}/{CVF} {Conference} on {Computer} {Vision}
  and {Pattern} {Recognition} (2021) 5008--5017.

\bibitem{kim_distilling_2021}
Y.~Kim, J.~Park, Y.~Jang, M.~Ali, T.-H. Oh, S.-H. Bae, Distilling {Global} and
  {Local} {Logits} {With} {Densely} {Connected} {Relations}, Proceedings of the
  {IEEE}/{CVF} {International} {Conference} on {Computer} {Vision} (2021)
  6290--6300.

\bibitem{kim_self-knowledge_2021}
K.~Kim, B.~Ji, D.~Yoon, S.~Hwang, Self-{Knowledge} {Distillation} {With}
  {Progressive} {Refinement} of {Targets}, Proceedings of the {IEEE}/{CVF}
  {International} {Conference} on {Computer} {Vision} (2021) 6567--6576.

\bibitem{szegedy_rethinking_2016}
C.~Szegedy, V.~Vanhoucke, S.~Ioffe, J.~Shlens, Z.~Wojna, Rethinking the
  {Inception} {Architecture} for {Computer} {Vision}, Proceedings of the {IEEE}
  {Conference} on {Computer} {Vision} and {Pattern} {Recognition} (2016)
  2818--2826.

\bibitem{reed_training_2015}
S.~Reed, H.~Lee, D.~Anguelov, C.~Szegedy, D.~Erhan, A.~Rabinovich, Training
  {Deep} {Neural} {Networks} on {Noisy} {Labels} with {Bootstrapping},
  arXiv:1412.6596 (2015).

\bibitem{xie_disturblabel_2016}
L.~Xie, J.~Wang, Z.~Wei, M.~Wang, Q.~Tian, {DisturbLabel}: {Regularizing} {CNN}
  on the {Loss} {Layer}, Proceedings of the {IEEE} {Conference} on {Computer}
  {Vision} and {Pattern} {Recognition} (2016) 4753--4762.

\bibitem{zhang_delving_2021}
C.~Zhang, P.~T. Jiang, Q.~Hou, Y.~Wei, Q.~Han, Z.~Li, M.-M. Cheng, Delving
  {Deep} {Into} {Label} {Smoothing}, IEEE Transactions on Image Processing 30
  (2021) 5984--5996.

\bibitem{simonyan_very_2014}
K.~Simonyan, A.~Zisserman, Very deep convolutional networks for large-scale
  image recognition, arXiv:1409.1556 (2014).

\bibitem{paszke_pytorch_2019}
A.~Paszke, S.~Gross, F.~Massa, A.~Lerer, J.~Bradbury, G.~Chanan, T.~Killeen,
  Z.~Lin, N.~Gimelshein, L.~Antiga, A.~Desmaison, A.~Kopf, E.~Yang, Z.~DeVito,
  M.~Raison, A.~Tejani, S.~Chilamkurthy, B.~Steiner, L.~Fang, J.~Bai,
  S.~Chintala, {PyTorch}: {An} {Imperative} {Style}, {High}-{Performance}
  {Deep} {Learning} {Library}, Advances in {Neural} {Information} {Processing}
  {Systems} 32 (2019).

\end{thebibliography}
\end{document}